\documentclass{article}
\usepackage{iclr2016_workshop,times}
\usepackage{times}
\usepackage{epsfig}
\usepackage{graphicx}
\usepackage{amsmath}
\usepackage{amssymb}
\usepackage{caption}
\usepackage{subcaption}
\usepackage{comment}
\usepackage{gensymb}
\usepackage{epstopdf}
\newcommand{\ignore}[1]{}
\DeclareRobustCommand\onedot{\futurelet\@let@token\@onedot}
\def\@onedot{\ifx\@let@token.\else.\null\fi\xspace}
\def\eg{\emph{e.g.}} 

\def\ie{\emph{i.e.}}

\usepackage[toc,page]{appendix}

 \usepackage{titlesec}

\titlespacing\section{0pt}{5pt plus 1pt minus 1pt}{0pt plus 1pt minus 1pt}

\titlespacing\subsection{0pt}{4pt plus 1pt minus 2pt}{0pt plus 1pt minus 1pt}

\titlespacing\subsubsection{0pt}{4pt plus 1pt minus 1pt}{0pt plus 1pt minus 1pt}

\usepackage[pagebackref=true,breaklinks=true,letterpaper=true,colorlinks,bookmarks=false]{hyperref}

\begin{document}
\title{Convolutional Models for Joint Object Categorization and Pose Estimation}

\author{    
\,\,\,\,\,\,\,\,\,\,\,\,Mohamed Elhoseiny\thanks{ Equal contribution}\\
{\tt\small m.elhoseiny@cs.rutgers.edu}
\and
\textbf{Tarek El-Gaaly$^{*}$}\\ 
{\tt\small tgaaly@cs.rutgers.edu}
\and
\textbf{Amr Bakry$^{*}$} \\
{\tt\small amrbakry@cs.rutgers.edu}
\and
\textbf{Ahmed Elgammal}\\
{\tt\small elgammal@cs.rutgers.edu}
\and
$\,\,\,\,\,\,\,\,\,\,\,\,\,\,\,\,\,\,\,\,\,\,\,\,\,\,\,\,\,\,\,\,\,\,\,\,\,\,\,\,\,\,\,\,\,\,\,\,\,\,\,\,\,\,\,\,\,\,\,\,$Computer Science Department, Rutgers University\\
}

\maketitle

\begin{abstract}
In the task of Object Recognition, there exists a dichotomy between the categorization of objects and estimating object pose, where the former necessitates a view-invariant representation, while the latter requires a representation capable of capturing pose information over different categories of objects. With the rise of deep architectures, the prime focus has been on object category recognition. Deep learning methods have achieved wide success in this task. In contrast, object pose regression using these approaches has received relatively much less attention. In this paper we show how deep architectures, specifically Convolutional Neural Networks (CNN), can be adapted to the task of simultaneous categorization and pose estimation of objects. We investigate and analyze the layers of various CNN models and extensively compare between them with the goal of discovering how the layers of distributed representations of CNNs represent object pose information and how this contradicts object category representations. We extensively experiment on two recent large and challenging multi-view datasets. Our models achieve better than state-of-the-art performance on both datasets.

\end{abstract}

\vspace{-2mm}
\section{Introduction}
\vspace{-2mm}
\label{S:Intro}


Impressive progress has been made over the last decade towards solving the problems of object categorization, localization and detection. It is desirable for a vision system to address two tasks under general object recognition: object categorization and object pose estimation (estimating the relative pose of an object with respect to a camera). Pose estimation is crucial in many applications. These two broad tasks are contradicting in nature. An optimal object categorization system should be able to recognize the category of an object, independent of its viewpoint. This means that the system should be able to learn viewpoint-invariant representations of object categories. In contrast, a pose estimation system requires a representation that preserves the geometric and visual features of the objects in order to distinguish its pose. 

This gives rise to a fundamental question: \emph{should categorization and pose estimation be solved simultaneously, and if so, can one aid the other?} Contrasting paradigms approach this question differently. Traditional instance-based 3D pose estimation approaches solve the instance-recognition and pose estimation problems simultaneously, given model bases of instances in 2D or 3D ~\citep{grimson1985recognition,lamdan1988geometric,lowe1987three,shimshoni1997finite}. Most recent object pose estimation approaches solve the problem within the detection process, where category-specific object detectors that encode part geometry are trained~\citep{Savarese07,Savarese08,mei2011robust,Payet2011,schels2012learning,pepik2012teaching}. Since part-geometry is a function of the pose, these approaches are able to provide coarse estimate of the object pose with the detection. However the underlying assumption here is that the categorization is done a-priori, and the representation is view-variant. Other recent approaches try to solve the pose estimation simultaneously with categorization through learning dual representations: view-invariant category representation and view-variant category-invariant representation~\citep{zhang_aaai_2013,bakry_untangling_2014}. 


With the rise of deep architectures, the main focus has been on category recognition. A wide success has been achieved on this task. In contrast, pose estimation has not received much attention. The impressive results of Convolutional Neural Networks (CNNs) in tasks of categorizations ~\citep{krizhevsky2012imagenet} and detection~\citep{Sermanet_EZMFL_2013, Malik_rcnn_2013} motivated many researchers to explore their applicability in different vision tasks. Several approaches recently showed successful results where they used networks that are pre-trained for a specific task (\eg categorization) and then use the representation of higher layers as features for another task ~\citep{NIPS13DeViSE,donahue2013decaf,fergus_visualizecnn_2013}. This process is known as transfer learning. As pointed out by~\citep{Yosinski_howtransferable_2014}, this process is useful when the target task has significantly smaller training data than what is needed to train the model. Typically the first $n$-layers are copied from the pre-trained network to initialize the corresponding layers for the target task. Within the CNN literature, typically the layers up until FC7 (which is the last layer before the output layer) are used for that purpose \citep{NIPS13DeViSE}.

Pose estimation is an example of a task that inherently suffers from lack of data. In fact the largest available dataset for multiview recognition and pose estimation has 51 object categories with a total of about 300 instances~\citep{lai2011large}. It is hard to imagine the availability of a dataset of thousands of objects where different views are sampled around each object in order to be able to train a learning machine such as a CNN with millions of parameters. Therefore, transfer learning is critical for this task. However the challenge lies in the contradicting objective that has been described in the first paragraph. Current CNN models are optimized for categorization, and therefore they are expected to achieve view invariant representation. Therefore it is not expected that feature representation at deeper layers are useful for pose estimation. However, feature representation in shallower layers tend to be more general and less category-specific and thus may hold enough information to discriminate between different poses. This is a key hypothesis that is explored in this paper and this work is the first exploration of the capability of CNNs on the task of object pose estimation.
%

The contributions of this paper are: (1) we show how CNNs can be adapted to the task of simultaneous categorization and pose estimation of objects, (2) we investigate how each of these tasks affect the other, \ie  how category-specific information can help estimate the pose of an object and how a balance between these contrasting tasks can be achieved, (3) we analyze different CNN models and extensively compare between them to find an efficient balance between accurate categorization and robust pose estimation, (4) we validate our work by extensive experiments on two recent large and challenging multi-view datasets. We achieve better than state-of-the-art performance on both datasets.


\vspace{-2mm}
\section{Related Work}
\vspace{-2mm}
\label{S:RelWork}

Due to the surge of work in deep architectures over the last few years, there has amassed a large number of research studies. Despite this, using CNNs for regression and capturing pose information is still a relatively unexplored area. This motivates the goals of this paper.

We focus on the most relevant work, in particular, studies that focus on understanding the functions of CNN layers and CNNs that solve for pose information. We also briefly touch upon previous approaches in object categorization and pose estimation. 

Although fundamentally different to object pose estimation, some research has explored using CNNs to recognize human pose \citep{Szegedy_deeppose_2013,LI_2014_CVPR_Workshops,pfister2014deep}. Recently~\citep{Gkioxari_rcnnactionpose_2014} proposed joint optimization on human pose and activity. In human pose estimation, there is no problem getting millions of image of people at different postures. Human activities are correlated with human poses, while in the object-pose domain the category is independent of pose. This makes joint learning of category and pose more challenging than joint learning of human activity and pose. Some  very recent work has explored joint detection and pose estimation using CNNs \citep{Tulsiani2015_viewpoints_keypoints}.

Recent in-depth studies explore the intricacies of CNNs; including the effects of transfer-learning and fine-tuning, properties and dimensions of CNN layers and the study of invariances captured in CNN layers (\eg \citep{Yosinski_howtransferable_2014,fergus_visualizecnn_2013,zisserman_devilindetails_2014}).
A data-centric analysis of existing CNN models for object detection has appeared in \citep{Pepik2015_holdingbackconvnets}.

A comprehensive review of recent work in object recognition and pose estimation is detailed in \citep{savarese_FeiFei10}. We highlight the most relevant research. 
Successful work have been done in estimating the object pose of a single object~\citep{Cyr2003,mei2011robust,schels2012learning}. This model, referred to as single-instance 3D model, has the limitation of being category-specific and does not scalable to a large number of categories and deal with high intra-class variation. Recently, detection and pose have been solved simultaneously (\eg  \citep{Tulsiani2015_viewpoints_keypoints,Pepik2015_3ddetectioninwild,xiang_pascal3d_2014,Savarese07,Savarese08,mei2011robust,Payet2011,schels2012learning,pepik2012teaching,optree}).  Most of these methods belongs to the category of limited-pose (discrete-pose) approaches since it uses classification for pose estimation. Very few works formulate the pose estimation problem as regression over a continuous space. In \citep{optree}, an object pose tree is built for doing multi-level inference. This  involves a classification strategy for pose which results in coarse estimates and does not utilize information present in the continuous distribution of descriptor spaces. Work presented in \citep{ZhangAAAI13} and \citep{Torki11} explicitly model the continuous pose variations of objects but the scalability of these models is limited. A more recent work \citep{bakry_untangling_2014} proposes a feedforward approach to solve the two problems jointly by balancing between continuous and discrete modeling of pose in order to increase performance and scalability. In these models, the nonlinearity in the category representations is not modeled, which is mandatory for many applications.

\vspace{-2mm}
\section{Motivation}
\vspace{-2mm}
\label{S:Motivation}

The first question we pose in this paper is \textit{how good are pre-trained representations of different CNN layers, without fine-tuning, for the task of pose estimation?} To answer this we analyzed a state-of-the-art CNN trained on ImageNet \citep{krizhevsky2012imagenet} by testing it on dense multi-view images from the RGBD dataset \citep{lai2011large} to see how well it represented object view-manifolds and hence able to estimate object poses. This CNN is composed of 8 layers: Conv1, Pool1, Conv2, Pool2, Conv3, Conv4, Conv5, Pool5, FC6, FC7, FC8. Pool indicates Max-Pooling layers, Conv indicates convolutional layers and FC indicates fully connected layers. 

\begin{figure}[hb!]
\centering
    \includegraphics[width=0.47\columnwidth, height=1.5in]{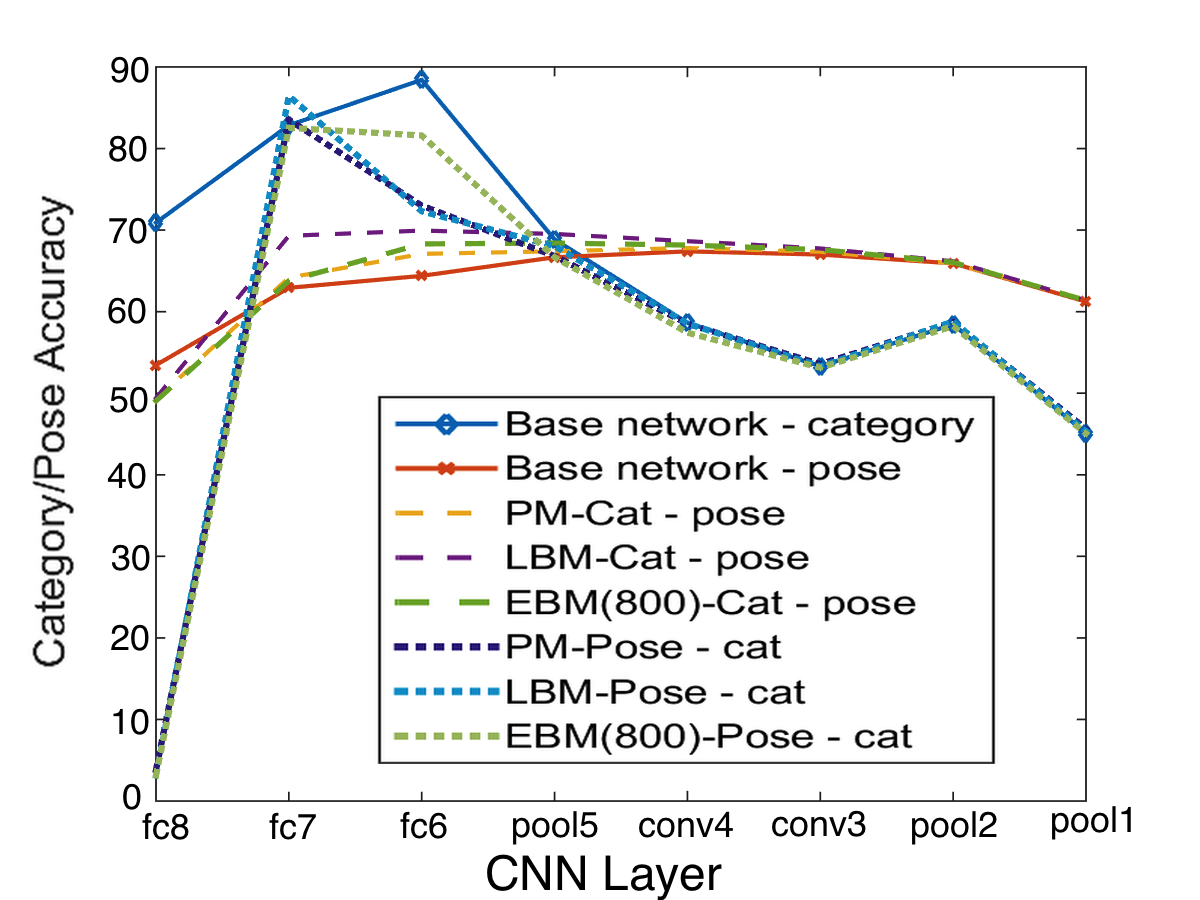}
   \includegraphics[width=0.47\columnwidth, height=1.5in]{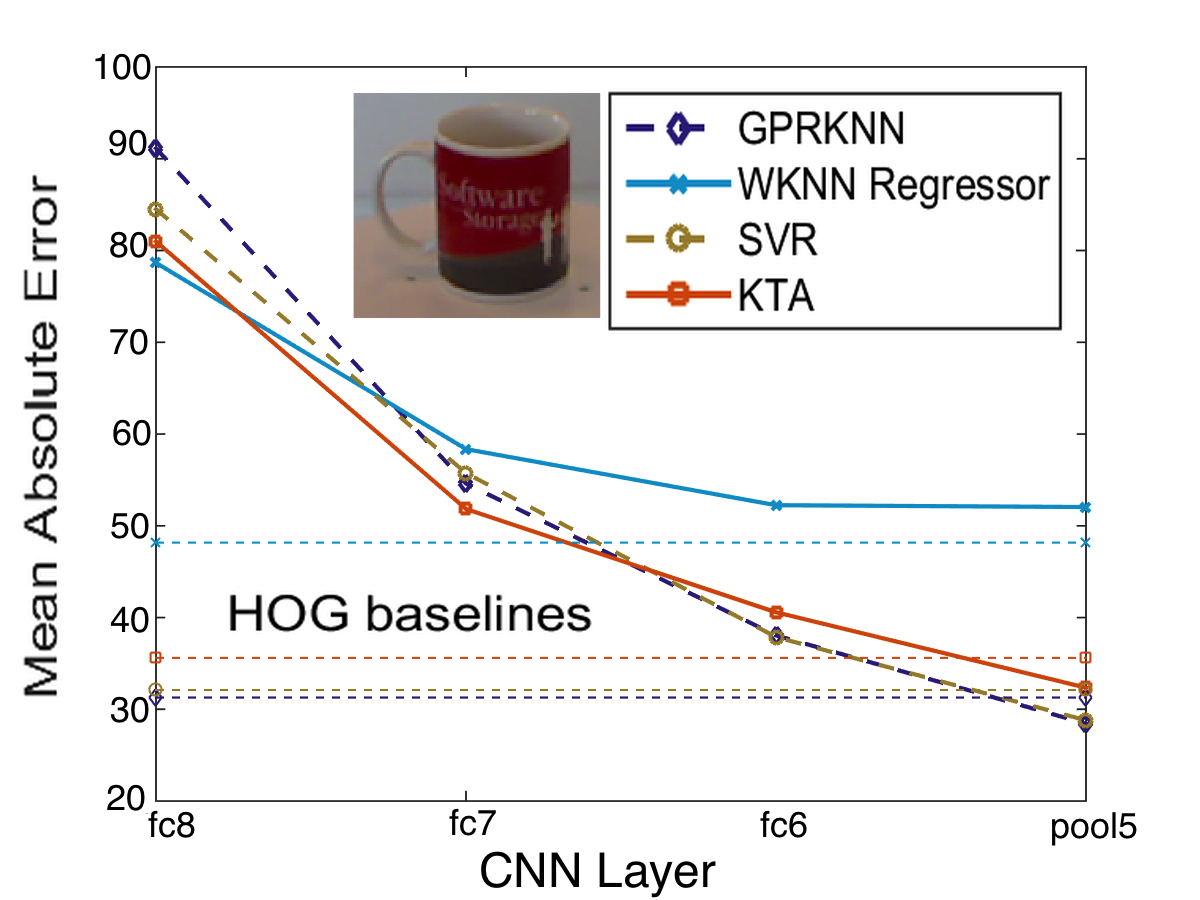}
\caption{\footnotesize  Left: Linear SVM categorization and pose regression performance based on feature encoding of different layers of a pre-trained CNN over all objects. The dotted lines are for cross-evaluation: for PM-Cat, LBM-Cat and EBM(800)-Cat represent the models' category representations evaluated on the task of pose estimation (to observe the effect of how category representations encode pose information). PM-Pose, LBM-Pose and EBM(800)-Pose are evaluated on the task of categorization to see how well pose representations in the CNN encodes categories. This is to show the complete pose-invariant representations of the layers when learning to categorize. Right: Pose regression on a single object - showing the Mean Absolute Error (in degrees) of various regression algorithms from FC8 to Pool5. The horizontal lines represent the regression performance on HOG feature descriptors computed on the images.}
\label{F:CatPoseModel0x}
\end{figure}

In order to quantitatively evaluate the representations of pose within the CNN, we trained both a pose regressor (using Kernel Ridge-Regression) and an SVM classifier for categorization (linear one-vs-all) on the features extracted from each of the layers. Fig.~\ref{F:CatPoseModel0x}-Left is the result of the regressor and classifier. It clearly shows the conflict in representation of the pre-trained network. For pose estimation, the performance increases until around Pool5 and then decreases. This confirms that shallow layers that have sufficient abstractive representation offer better feature encoding for pose estimation. It appears that Pool5 provides a representation that captures the best compromise in performance, between categorization and pose discrimination.

In Fig~\ref{F:CatPoseModel0x}-Left we report cross-evaluation of categorization using pose features and vice versa. FC8 (output) which is task specific, does not perform good pose estimation, while FC6/FC7 perform much better. It is interesting to observe the opposite is not true; when optimizing on pose, much of the category-specific information is still represented by the features of the CNN, as seen by the increase in performance of category recognition using the pose-optimized features. 

We further explored using other regressors on multiple views of a single object instance (GPR~\citep{book_GP}, WKNN~\citep{Kandola_spectralkernel_2001}, SVR~\citep{Kandola_spectralkernel_2001}, KTA~\citep{Kandola_spectralkernel_2001}). We use a coffee mug instance that has enough visual and shape features to discriminate its poses. Fig~\ref{F:CatPoseModel0x}-Right shows the MAE of the pose regression. The results confirm that the pose representation improves as we approach Pool5. This indicates that Pool5 has the best representation of the object's view-manifold. We also found that the performance of features based on Pool5 are the closest to correlate with the performance when using HOG features on the objects' multi-view images (Fig~\ref{F:CatPoseModel0x}-Right). This further proves that Pool5 has the abstraction capability to represent pose efficiently. 

It is important to point out here that, in addition to our analysis, in-depth manifold analysis was previously conducted to analyze the object-view manifolds and their representations within CNN layers. This can be found in \citep{bakry2015digging}. This in-depth study corroborates the conclusions we make here.

\vspace{-2mm}
\section{Analyzed Models}
\vspace{-2mm}
\label{S:PropModels}

We used a state-of-the-art CNN built by  \cite{krizhevsky2012imagenet} as our baseline network in our experiments (winner of the LSVRC-2012 Imagenet Challenge \cite{li_imagenet_2014}). We refer to this model as \textit{Model0: base network}. This model is pre-trained on Imagenet. The last fully connected layer (FC8) is fed to a 1000-way softmax which produces a distribution over the category labels. Dropout was employed during training and Rectified Linear Units (ReLU) were used for faster training. Stochastic gradient descent is used for back propagation. Model0 is not fine-tuned, and thus an analysis of it shows how the layers of a CNN trained on categorization of ImageNet lacks the ability to represent pose efficiently. Throughout this study we vary the architecture of the base network and the loss functions. All other models described are pre-trained on ImageNet and fine-tuned on each of the two large dataset we experimented on.



We propose and investigate four different CNN models: Parallel Model (PM), Cross-Product Model (CPM), Late Branching Model (LBM) and Early Branching Model (EBM). 

\begin{figure*}
  \centering
  \includegraphics[width=4in, height=2.5in]{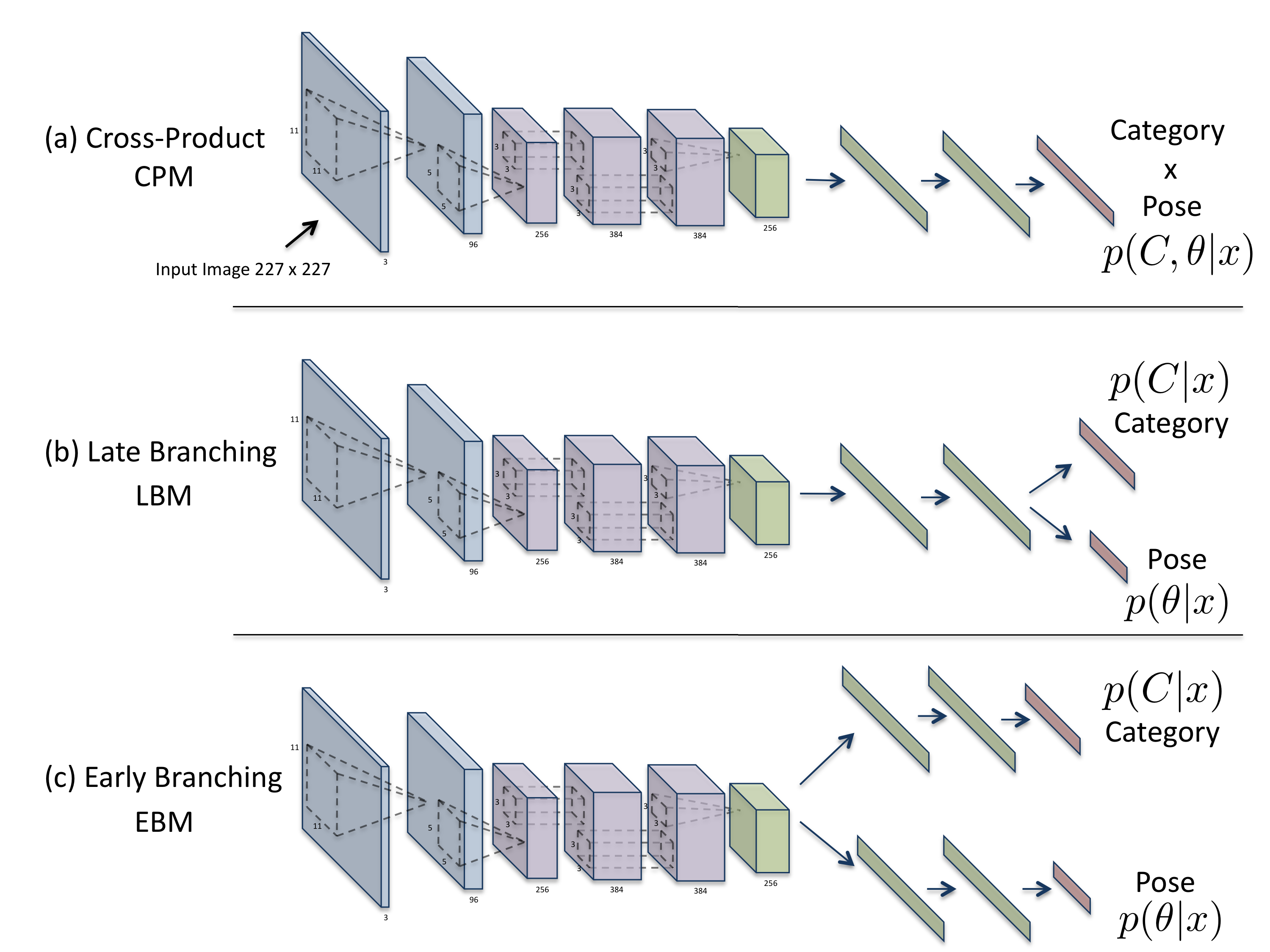}
  \caption{\footnotesize The proposed models showing the joint loss layer in CPM, late branching in LBM and early branching in EBM. Blue layers correspond to layers with convolution, pooling and normalization. Violet colored layers correspond to layers with just convolution. Green layers correspond to fully-connected layers.} 
  \label{F:Models}
\end{figure*}

PM is a parallel version of the base network; two parallel and independent versions of the base network, one for categorization and one for pose. CPM has an output layer with units for each category and pose combination to jointly train (depicted in Figure \ref{F:Models}-a). LBM and EBM models are also depicted in Figure \ref{F:Models}. LBM branches into two last layers, one for categorization and one for pose. EBM performs early branching into two subnetworks, each specialized in categorization and pose estimation, respectively. The output layer FC8 is not merged but instead the LBM and EBM networks are optimized over two loss functions, one concerned with building view-invariance representations for categorization and the other with category-invariant representations for pose estimation. Because of the branching, this causes two units to be active, one in each branch, at the same time. The only work that has done something similar to this is the work by\cite{Gkioxari_rcnnactionpose_2014}.

All losses are optimized by the multinomial logistic regression objective, similar to~\cite{krizhevsky2012imagenet} (\textit{Softmax} loss). We denote softmax loss of label $l\in\{l^c,l^p\}$ and image $x$ as $loss_i(x,l)$, where $i$ indicates if this loss is over category or pose modes, $l^p$ and $l^c$ are the labels for pose and category, respectively.



In the following subsections we describe each model in detail.

\textbf{Parallel Model (PM)}: This model consists of two base networks running in parallel, each solving categorization and pose estimation independently. There is no sharing of information between the two networks. 
The goal of this model is to see how well the traditional CNN is capable of representing object-view manifolds and hence estimating object pose, independent of category-specific information. The category and pose losses minimized in this model are: $loss_c(x,l^c)$ and $loss_p(x,l^p)$, one for each of the tasks of categorization and pose estimation, respectively. 

\textbf{Cross-Product Model (CPM)}: CPM explores a way to combine categorization and pose estimation by building a last layer capable of capturing both (see Fig\ref{F:Models}-a). We build a layer with the number of units equivalent to the number of combinations of category and pose, \ie the cross-product of category and pose labels. The number of categories varies according to the dataset as we will see. The pose angles (in this case yaw or azimuth angle of an object) is discretized into angle bins across the viewing circle. This is the case with all our pose estimating models. The loss function for CPM is the softmax loss over the cross-product of category and pose labels: $loss(x,l^p \times l^c )$. 

\textbf{Late Branching Model (LBM)}: We introduce a change in the architecture by splitting/branching the network into two last layers, each designed to be specific to the two tasks: categorization and pose estimation. Thus, this network has a shared representation for both category and pose information up until layer FC7 (see Fig\ref{F:Models}-b). 

The goal is to learn category and pose information simultaneously from the representations encoded in the previous  layers of the CNN. The question behind this model is whether or not one last layer would be enough to recover the pose information from the previous layers, in other words untangle the object view-manifold and give accurate pose estimates. In other words, one can think of this as testing the ability of the deep distributed representations of a CNN in holding both category-specific pose-invariant information as well as pose-variant information. LBM is trained using a linear combination of losses over category and pose: $\lambda_1 \cdot loss_c(x,l^c) + \lambda_2 \cdot loss_p(x,l^p)$ where $\lambda_1, \lambda_2$ are weights found empirically (see appendix~\ref{app_weights}).


\textbf{Early Branching Model (EBM)}: The question of moving the branching to an earlier layer in the network poses itself here: \textit{Can the branching be moved earlier in the network to where the pose knowledge is still well preserved and in fact maximal across the layers?} 

From our experiments (described later on) we observe that the objects' view-manifolds are maximally represented at Pool5. Thus, this network has a shared representation for both category and pose information up until layer Pool5. At Pool5 it branches out into two subnetworks, that are jointly optimized using a combined loss function (same as for LBM): $\lambda_1 \cdot loss_c(x,l^c) + \lambda_2 \cdot loss_p(x,l^p)$. Similar to LBM, it is important to note that this network optimizes over two losses. 
This model achieves the most efficient balance between categorization and pose estimation and achieve state-of-the-art results on two large challenging datasets, as we shall see in Section \ref{S:Expr}. 


%
%
\vspace{-2mm}
\section{Datasets}
\vspace{-2mm}
\label{S:Datasets}
\subsection{RGBD Dataset}
\label{ss:rgbd}

One of the largest and challenging multi-view datasets available is the RGB-D dataset \cite{lai2011large}. It consists of 300 tabletop object instances over 51 different categories. Images are captured of objects rotating on a turn-table, resulting in dense views of each object. The camera is positioned at three different heights 
with elevation angles: 30$^\circ$, 45$^\circ$ and 60$^\circ$.  

In previous approaches the middle height ($45^\circ$) is left out for testing \cite{optree,zhang_aaai_2013,bakry_untangling_2014,elgaaly_mkl}. This means that object instances at test time have been seen before from other heights during training. For this dataset it was important to experiment with an additional training-testing split of the data to give more meaningful results. We wanted to ensure that objects at test time had never seen before. We also wanted to make sure that the instances we are dealing with have non-degenerate view manifolds. We observed that many of the objects in the dataset are ill-posed, in the sense that the poses of the object are not distinct. This happens when the objects have no discriminating texture or shape to be able to identify its poses (\emph{e.g.} a texture-less ball or orange). This causes object-view manifold degeneracy. For this reason, we select 34 out of the 51 categories as objects that possess variation across the viewpoints, and thus are not ill-posed with respect to pose estimation. We split the data into training, validation and testing. In this datasets, most categories have few instances; therefore we left out two random instances per category, one for validation and one for testing. In the case where a category has less than 5 instances, we form the validation set for that category by randomly sampling one object instance from the training set. We also left out all the middle height for testing. Thus, the testing set is composed of unseen instances and unseen heights and this allows us to more accurately evaluate the capability of CNNs in discriminating categories and poses of tabletop objects. We call this split, Split 1. In order to compare with state-of-the-art we also used the split used by previous approaches (we call this Split 2). 

\subsection{Pascal3D+ Dataset}
We experiment on the recently released challenging dataset of multi-view images, called Pascal3D+ \cite{xiang_pascal3d_2014}. Pascal3D+ is very challenging because it consists of images \textit{in the wild}, in other words, images of object categories exhibiting high variability, captured under uncontrolled settings, in cluttered scenes and under many different poses. Pascal3D+ contains 12 categories of rigid objects selected from the PASCAL VOC 2012 dataset \cite{Pascal}. These objects are annotated with pose information (azimuth, elevation and distance to camera). Pascal3D+ also adds pose annotated images of these 12 categories from the ImageNet dataset \cite{li_imagenet_2014}. The \textit{bottle} category is omitted in state-of-the-art results. To be consistent, we do the same. This leaves 11 categories to experiment with. There are about 11,500 and 7,000 training images in ImageNet and Pascal3D+ subsets, respectively. We take a small portion of these images for validation and use the rest for training. For testing, there are about 11,200 and 6,900 testing images for ImageNet and Pascal3D+, respectively. On average there are about 3,000 object instances per category in Pascal3D+ captured \textit{in the wild}, making it a challenging dataset for estimating object pose.


%
%
%
\vspace{-2mm}
\section{CNN Layer Analysis}
\begin{figure}
\begin{minipage}{1.0\linewidth}
\centering
	\includegraphics[width=0.49\columnwidth]{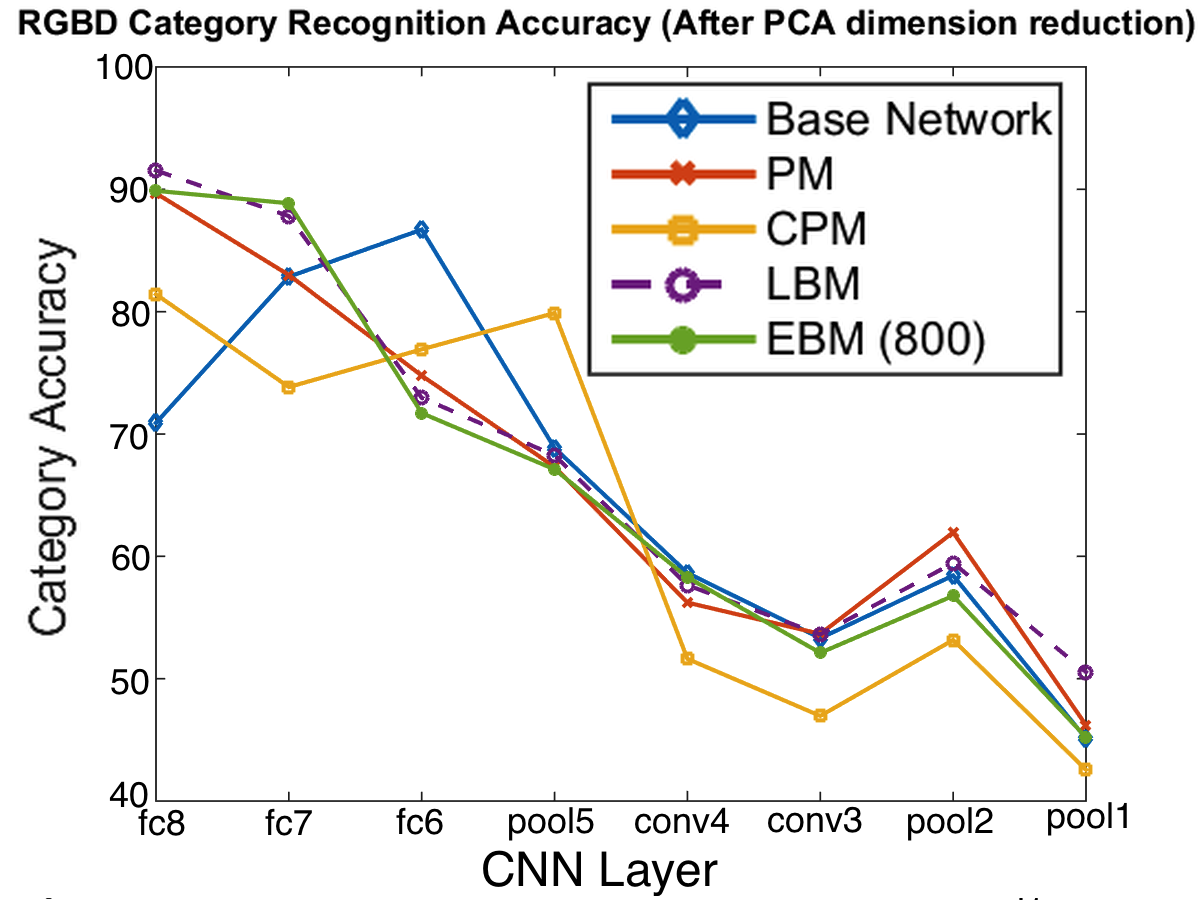}
     \includegraphics[width=0.49\columnwidth]{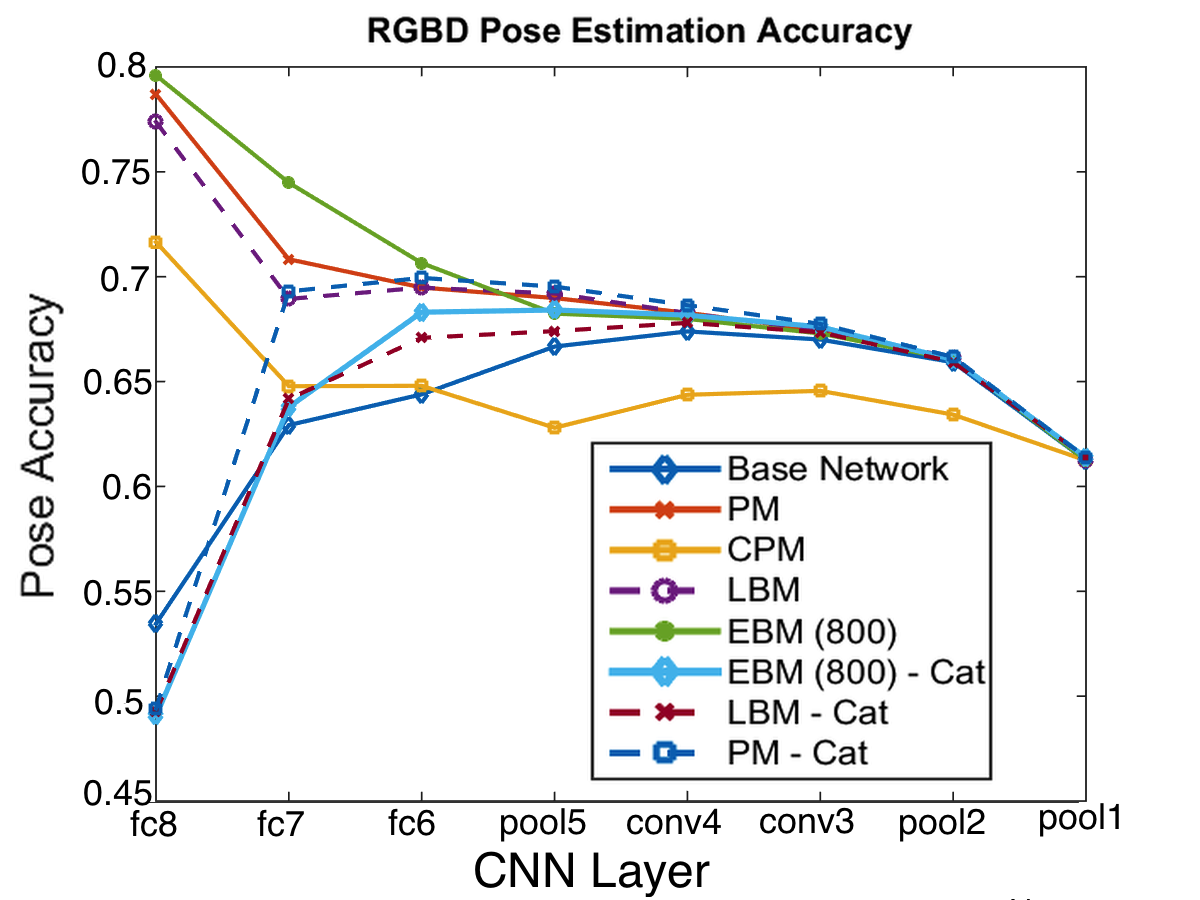}
       \caption{  Analysis of layers trained on the RGBD dataset. Left: the performance of linear SVM category classification over the layers of different model. Right: the performance of pose regression over the layers of different models (including the category parts of some of the models - this shows the lack of pose information encoded within the object category representations)}
    \label{fig:SVMRidge_rgbd}
    
       \end{minipage}
\begin{minipage}{1.0\linewidth}
\centering
	\includegraphics[width=0.49\columnwidth]{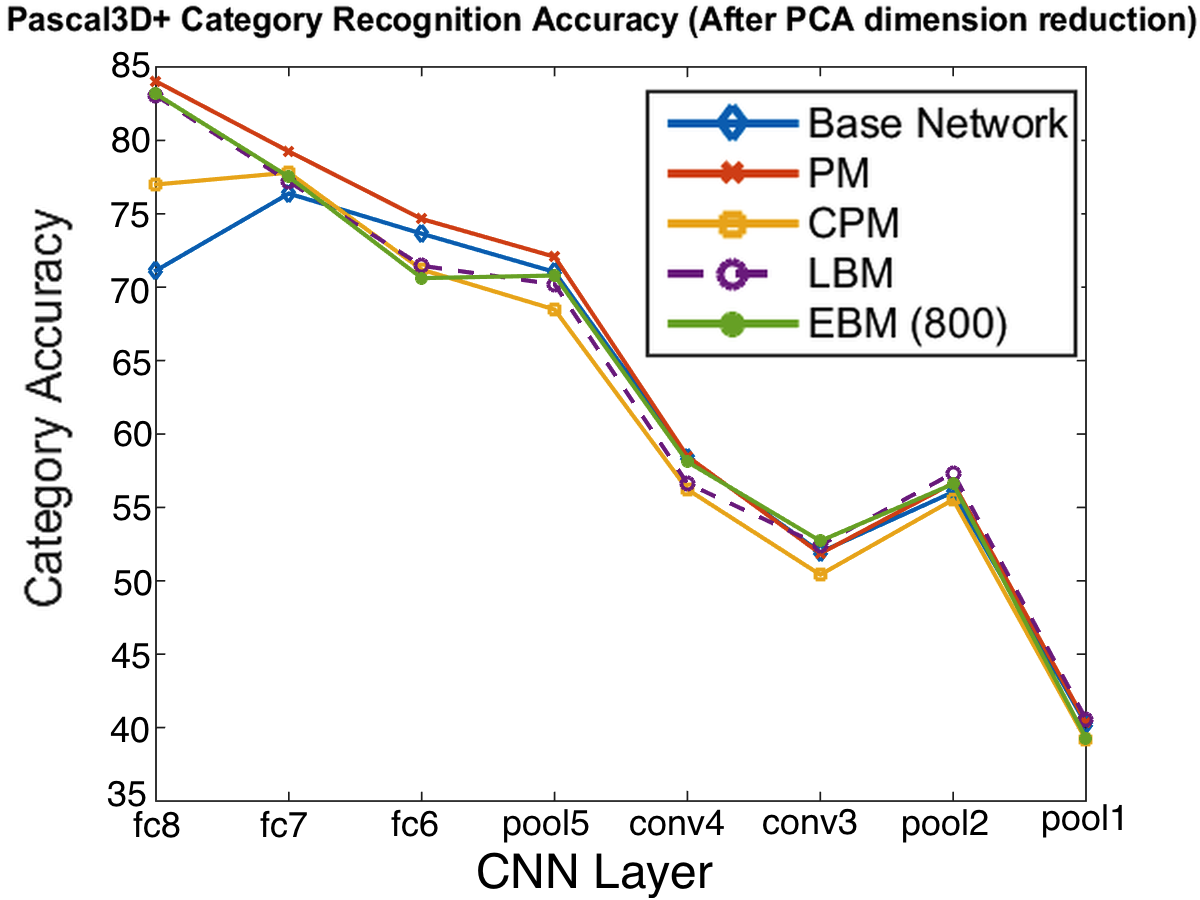}
     \includegraphics[width=0.49\columnwidth]{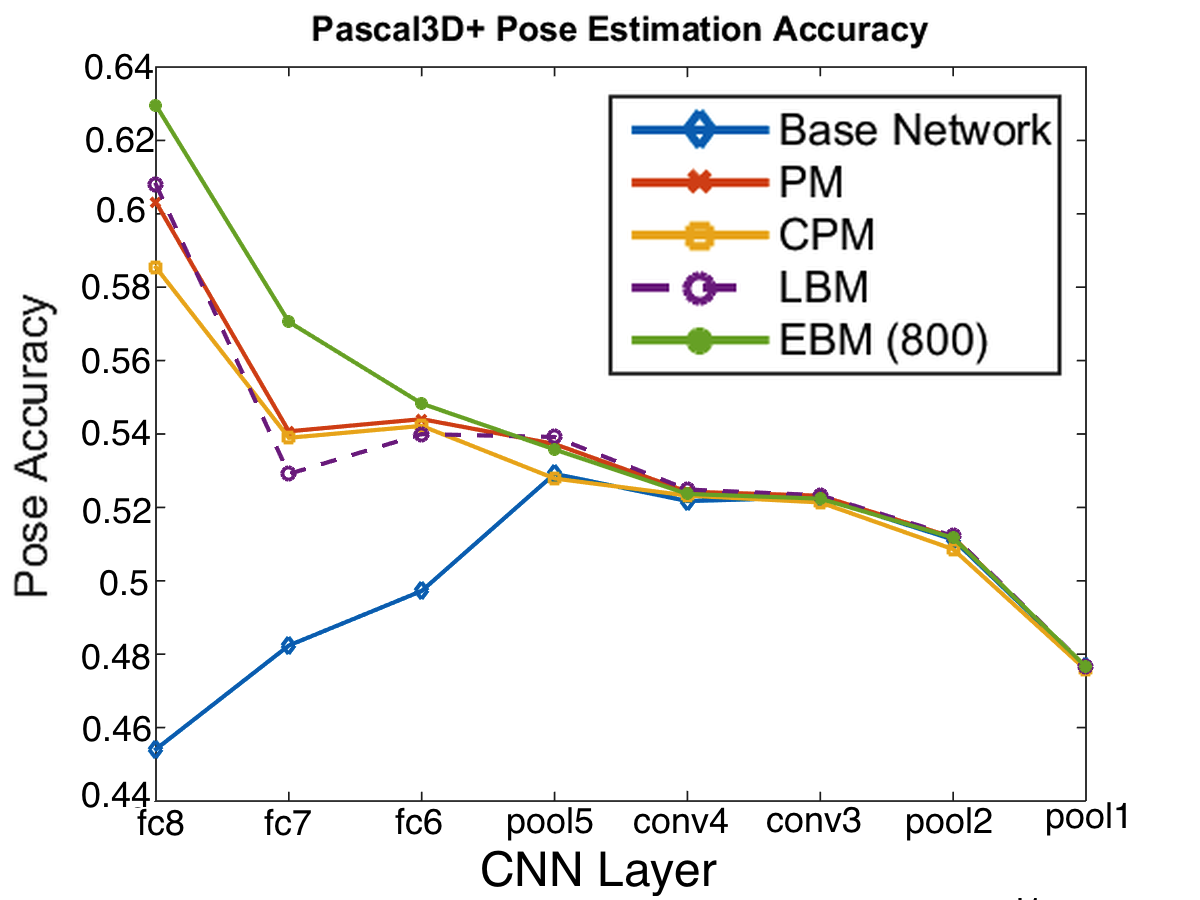}
	    \caption{  Analysis of layers trained on the Pascal3D+ dataset. Left: the performance of linear SVM category classification over the layers of different model. Right: the performance of pose regression over the layers of different models}\label{fig:SVMRidge_pascal}
	     \end{minipage}
\end{figure}

\vspace{-2mm}
\label{S:analysis}
Similar to the analysis performed in Section \ref{S:Motivation}, we do the same on all our described models. This gives insight into the ability of these models to represent pose and the intrinsic differences between them. We perform kernel Ridge-regression and SVM classification on each layer of the CNN models. The results of this analysis on the two muti-view datasets are shown in Fig. \ref{fig:SVMRidge_rgbd} and \ref{fig:SVMRidge_pascal}.
%

From Fig. \ref{fig:SVMRidge_rgbd} and \ref{fig:SVMRidge_pascal}, we can see that the base network monotonically decreases in pose accuracy after layer Pool5. Pool5 seems to again hold substantial pose information, before it is lost in the following layers. This is the premise behind the design of our EBM model. EBM is able to efficiently untangle the object-view manifold and achieve good pose estimation on the branch specific to pose estimation. 

%


From Fig. \ref{fig:SVMRidge_rgbd} and \ref{fig:SVMRidge_pascal}, it can be observed that the LBM is able to achieve a good boost in pose performance at its last layer. From \ref{fig:SVMRidge_rgbd}-right, it can be observed that at the layers Conv4 and Pool5 EBM has slightly worse accuracy than LBM and PM on the RGBD dataset. This indicates that the optimization is putting emphasis on the category information just before branching to achieve better pose estimation at deeper layers.

CPM does quite  worse than the other models on both datasets, in both categorization and pose estimation. This can be seen in Fig. \ref{fig:SVMRidge_rgbd}-Left and \ref{fig:SVMRidge_rgbd}-Right and to some extent in Fig. \ref{fig:SVMRidge_pascal}. The reason for this lies in the fact that CPM shares information to jointly optimize over category and pose. The drop is more evident in the task of categorization, indicating again that category information aids in estimating the pose, but not the other way round. The drop is more on the RGBD dataset because there are a lot more categories than Pascal3D+ and thus a lot more inter-class confusion. This is analogous to using category labels to separate between objects of different categories which may help bring similar posed objects of the same category together in the \textit{latent} space encoded in the layers. On the other hand there is no clear untangling of the object-view manifold, where the pose information is stored, and thus this lack of pose information negatively impacts the categorization of objects.

\begin{figure}[t!]
\centering
 \includegraphics[width=0.95\columnwidth]{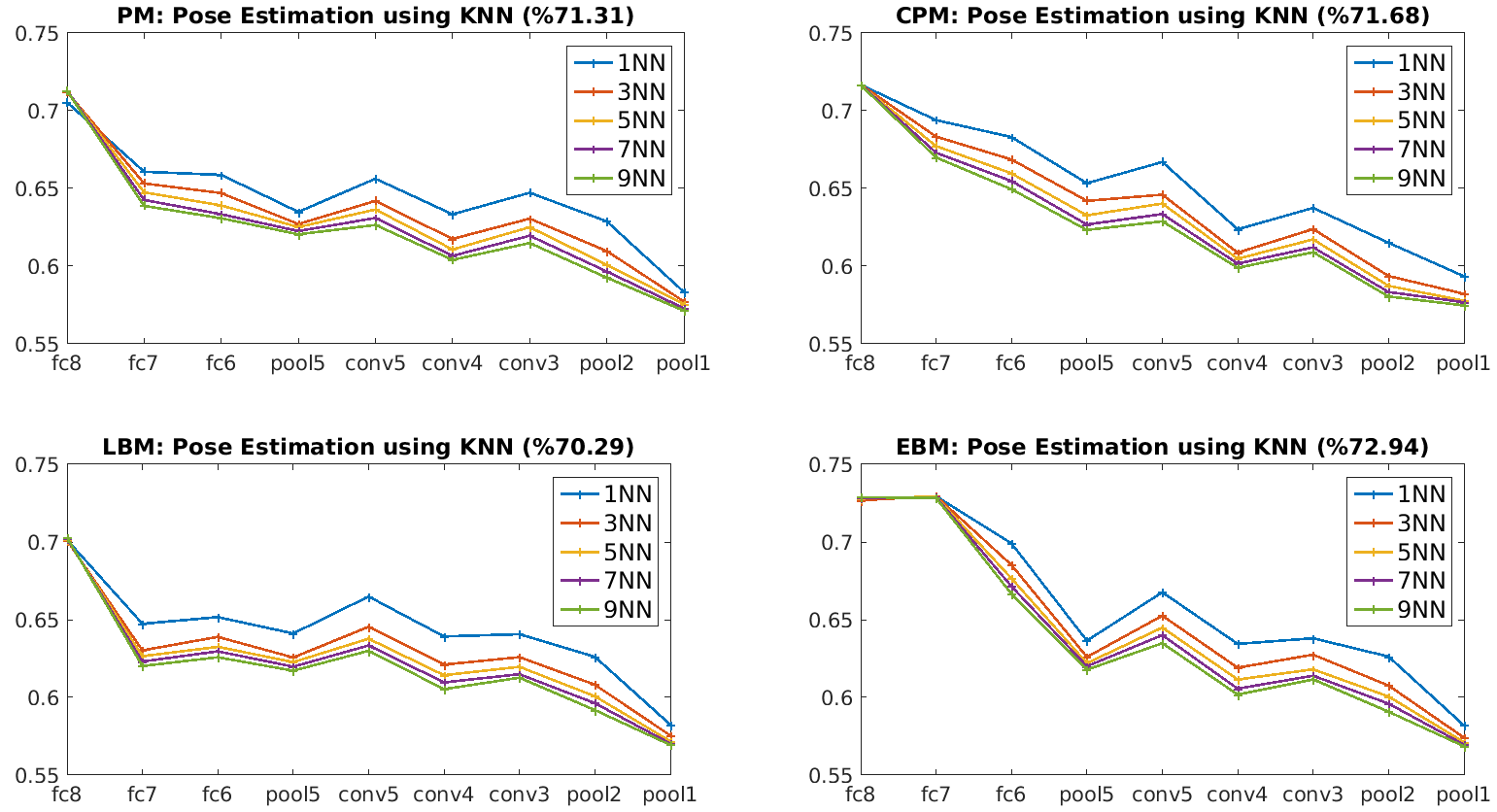}
   \caption{\small Comparison between EBM and LBM for categorization and pose estimation at each layer of the CNN using $k$-NN with varying $k=\{1,3,5,7,9\}$. This experiment was conducted on the Pascal3D dataset.} 
  \label{F:kNN}
\end{figure}

\textbf{$k$-NN Layer Analysis}
We conduct $k-Nearest Neighbor$ pose estimation over the Pascal3D dataset on all the layers of the 4 models with varying neighborhood sizes (shown in Fig. \ref{F:kNN}). 
Comparing the two models (LBM and EBM), we gain slight improvement in categorization and a large improvement in pose estimation performance when using EBM. From Figure~\ref{F:kNN} and \ref{F:kNN2} in the Appendix, we conclude that as we go deeper into the network - up to layer Conv5 - we gain more category separation and object-view manifold preservation. This shows how the early branching better resolves the contradiction between the pose estimation and categorization tasks while sharing the low level filter representations that are helpful for both tasks. After Conv5, there are two common layers in EBM. In these two layers, linear separability between categories increases (seen in Fig. ~\ref{F:kNN2}), but the object-view manifolds collapse (as seen in Fig. \ref{F:kNN}). This hurts the pose estimation. At the same time, this supports the aforementioned claim that enforcing better categorization (fine-tuning) hurts pose estimation. In our best performing model (EBM), in Figure~\ref{F:kNN}, remarkable improvement to the pose object-view manifold is attained. For pose, the drop in KNN-classifier as the K increases vanishes when going deeper in network; see FC6 and FC7 layers EBM in Fig.~\ref{F:kNN}. KNN figure for categorization on Pascal3D dataset could be seen in appendix A. In a similar behavior EBM behaves better than CPM and PM. An interesting behavior that CPM works clearly on Pascal3D dataset compared to RGBD; see Appendix~\ref{app_KNN1} for KNN analysis on RGBD dataset. This is due that RGBD dataset has both dense poses and also  larger number of categories (~ 5 times Pascal3D). This increases the information/uncertainty to model that are beyond the capacity of CPM for RGBD dataset and generally as the number of categories and poses increase.   

\textbf{Local Pose Measurement Analysis: } 
In Appendix~\ref{sec:localmeas}, we further analyzed  four  local measurement pose analysis proposed in~\citep{bakry2015digging} to analyze layers  of the five models we study against the pose manifold. The purpose of this analysis is to show how the learning representations for each model is untangle to the circle manifold where the pose inhabits 


\vspace{-2mm}
\section{Experiments}
\vspace{-2mm}
\label{S:Expr}


Here we describe the experimental setup and present the quantitative results of our experiments as well as comparisons with state-of-the-art.

\subsection{Training and Testing}

All models are trained by back propagation with Stochastic gradient descent. Refer to  appendix~\ref{app_training} for parameter settings, \emph{e.g.} learning rate, decay, \emph{etc}. At training time, we randomly sample 227x227 patches from the down-scaled 256x256 images. At test time the center 227x227 patches are taken.

All classification losses are optimized by the multinomial logistic regression objective. Similar to~\citep{krizhevsky2012imagenet}, we optimized it by maximizing the average of the log-probability of the correct label under the prediction distribution across training cases. The pose softmax output (FC8) layer produces the pose probability distribution given the image. For each of the category and pose losses, the gradient with respect to the CNN parameters is computed which is then fed into CNN training for back propagation.

All the results that are presented in the paper were based on the prediction of $argmax_{pose} p(pose|x)$, where $pose$ is one of the 16 pose bins and $x$ is the given image. 

In addition, we conduct an experiment where we predict the pose by computing the expected pose in the distribution of $p(pose|x)$; see Eq.~\ref{eqint}. 
\begin{equation}
\small
E(p(pose|x)) =  \sum_i   p(pose_i|x) \times \phi(pose_i)
\label{eqint}
\end{equation}where $\phi(pose_i)$ is the center angle of the corresponding bin $pose_i$ (the pose of the $i^{th}$ bin). 


The detailed definitions of the performance metrics used in our experiments are described in appendix~\ref{app_metrics}.

\subsection{Results}

Table \ref{T:CompCNN} shows the category recognition and pose estimation performance for the different models on the two training-testing splits of the RGBD dataset. Table  \ref{T:CompSOA} shows our best performing model EBM compared to state-of-the-art approaches. Using the pose prediction rule of eq. \ref{eqint}, the pose accuracy of EBM(800) increased from 78.83\% to 79.30\% for the $argmax$ prediction on split 2 of RGBD. 

Looking at the closest previous approach in Table \ref{T:CompSOA},~\citep{bakry_untangling_2014} achieves 96.01\% classification accuracy. This is achieved using both visual and depth channels. We only used RGB (without depth) in our approach. \citep{bakry_untangling_2014} achieves a lower 94.84\% with RGB only, which shows the advantage of our CNN models for classification. 

We achieve 2.3\% increase in category recognition and about 2\% increase in pose estimation (79.30\%) using EBM(800), when compared with state-of-the-art. These measurements are likely to increase further when using EBM(4096), as we see slight improvement of EBM(4096) over EBM(800) in Table \ref{T:CompSOA}. It is also possible that running $k$-NN on top of the layer features could improve performance further. We achieved 97.14\% categorization using EBM. We also achieved 99.0\% classification accuracy using Nearest Neighbor classification on the Pool5 layer of EBM, which indicates that we learned better convolutional filters.

Table \ref{T:CompSOAPascal} shows the performance of our models on Pascal3D+. We compare the accuracy achieved by our models with state-of-the-art results by \citep{zhang2015factorization,xiang_pascal3d_2014,Pepik2015_3ddetectioninwild,Tulsiani2015_viewpoints_keypoints}. It must be noted here that we are solving slightly different problems to some of these approaches. In \cite{xiang_pascal3d_2014}, the authors solve detection and pose estimation, assuming correct detection. On the other hand \citep{zhang2015factorization} solve just pose estimation, assuming that the object categories are known. \citep{Pepik2015_3ddetectioninwild,Tulsiani2015_viewpoints_keypoints} solve joint detection and pose estimation. In our case we are jointly solving both category recognition and pose estimation, which can be considered a harder problem than that of \citep{zhang2015factorization} and \citep{xiang_pascal3d_2014}. Our pose estimation performance is better than all these previous approaches. For the sake of this comparison, we computed the pose performance using the metrics applied in~\citep{zhang2015factorization}. These metrics are pose accuracy for images with pose errors $<22.5^\circ$ and $<45^\circ$. 


Table~\ref{T:CompSOAPascal} shows both our categorization and pose estimation results on Pascal3D compared against previous approaches. The table indicates 13.69\% improvement of our method over ~\citep{zhang2015factorization} (the best performing previous approach) in $pose <22.5^\circ$ metric and $4\%$ improvement in $pose <45^\circ$, which are significant results. It is important to note that comparing to \citep{xiang_pascal3d_2014,Tulsiani2015_viewpoints_keypoints,Pepik2015_3ddetectioninwild} is slightly unfair because these works solve for detection and pose simultaneously, while we do not solve detection.

We also show our performance when including ImageNet images in the training set and also the test set - see Table~\ref{T:CompSOAPascal} (rows 7-8). The results show the benefit of ImageNet training images which boosts pose performance to $76.9\%$ (from $57.89\%$) and $88.26\%$ (from $63.0\%$) for $pose <22.5^\circ$ and $pose <45^\circ$, respectively. 


In Table \ref{T:CompSOAPascal}, on the \textit{in the wild} images of Pascal3D+, our EBM model achieves an impressive increase of $\sim$8\% and $\sim$5\% over the state-of-the-art models using the two pose accuracy metrics, respectively. 



\begin{center}
\begin{table}
\centering
\caption{\small A summary of all the results of the CNN models. Split 2 is the traditional RGBD dataset split. Split 1 is the one we describe that better evaluates our experiments. Split 2 is the state-of-the-art training-testing split. C indicates category performance and P indicates pose accuracy (where it is measured using 3 different metrics consistent with state-of-the-art. P ($<22.5^\circ$) indicates that the pose accuracy is measured for objects where the pose error was less than $22.5^\circ$}
\scalebox{0.9}{
\scriptsize 
    \begin{tabular}{ | p{5cm} | p{0.4cm} | p{0.6cm} | p{2cm} | p{2cm} |p{1.0cm} |p{1.0cm} |}
    \hline
    Model & Split & C\% & P\%  ($<22.5^\circ$) & P\% ($<45^\circ$) &  P\% (AAAI)\\ \hline
   PM  & 1 & 89.63 & 69.58 & 81.09 &81.21  \\ \hline
    CPM  & 1 & 80.68 & 63.46& 75.45 & 77.35 \\ \hline
    LBM & 1 & \textbf{91.48} & 68.25& 79.31& 79.94 \\ \hline
    EBM (4096) & 1 & 89.94 & \textbf{71.49} & 82.19 & \textbf{82.00} \\
     EBM (800) & 1 & 89.84&  71.29 & \textbf{82.29} & {81.91} \\ 
      EBM (400) & 1 & 89.77 & 70.80& 81.73 & {81.65} \\ 
        EBM (200) & 1 & 90.11 &67.70 & 79.43 & {79.77} \\ 
           EBM (100) & 1 & 90.34 &  69.15&  80.09& {80.36} \\ 
     \hline \hline
    EBM (800) & 2 & \textbf{97.14} & \textbf{66.13 }& \textbf{77.02} & \textbf{78.83} \\ \hline 
EBM (4096) & 2 & {97.07} & 65.82 & 76.51 & {78.66} \\ \hline    
    \hline
    SVM/Kernel Reg - Model 0 (best category - FC6) & 1 & 86.71 & - & - & 64.39 \\ \hline SVM/Kernel Reg - Model 0 (best pose - Conv4) & 1 & 58.64 & - & - & 67.39 \\ \hline
     SVM/Kernel Reg - HOG & 1 & 80.26 & - & - & 27.95 \\ \hline 
    \end{tabular}
    }
       \label{T:CompCNN}
    \end{table}
\end{center}
\begin{center}
\begin{table}
\centering
\scriptsize
    \caption{ \small RGBD Dataset: Comparison with state-of-the-art approaches on category recognition and pose estimation (Ours use only RGB channel).}
\scalebox{0.9}{
    \begin{tabular}{ | p{3cm} | p{5cm} | p{3cm} |}
    \hline
    Approach & Category \% & Pose (AAAI) \% \\ \hline
    \citep{optree} & 94.30 (RGB + Depth) & 53.50  \\ \hline
    \citep{bakry_untangling_2014} & 94.84 (RGB only)/ 96.01 (RGB+ Depth) & 76.01 \\ \hline 
    \citep{zhang_aaai_2013} & 92.00 (RGB only)/ 93.10 (RGB + Depth) & 61.57 \\ \hline
    \citep{bakry_wacv_2016} & 85.00  & 77.31 \\ \hline 
    Ours (EBM(800)) & \textbf{97.14}  & \textbf{79.30}  \\ \hline
    \end{tabular}
    }
    \label{T:CompSOA}
    \end{table}
\end{center}
\begin{table}
\vspace{-4mm}
\centering
    \caption{Pascal3D dataset~\citep{xiang_pascal3d_2014}: Comparison with state-of-the-art approaches on category recognition and pose estimation. The AAAI pose metric is the performance metric used in\citep{zhang2015factorization,optree,zhang_aaai_2013}.}
\scriptsize
    \begin{tabular}{ | p{2cm} | p{1.5cm} | p{3cm} |p{3cm} |p{2cm}|}
    \hline
    \multicolumn{5}{|c|}{Train: Pascal12, Test: Pascal12} \\
    \hline
    Approach & Category & Pose \% (error $< 22.5$)&  Pose \% (error $< 45$)& Pose AAAI metric\\ \hline    
    \citep{xiang_pascal3d_2014} & {-} &  {15.60} &  {18.70} & {-} \\ \hline 
    \citep{pepik2012teaching} & {-} &  {17.30} &   {21.50} & {-} \\ \hline 
        \citep{Pepik2015_3ddetectioninwild} & {-} &{18.60} &  {27.60}   & {-} \\ \hline 
            \citep{Tulsiani2015_viewpoints_keypoints} & {-} & {36.00}  &  {44.50}  & {-} \\ \hline 
    \citep{zhang2015factorization} & - & 44.20 &  59.00 & - \\ \hline 
      EBM (4096) & 83.0 & \textbf{51.80} & \textbf{64.27}  & \textbf{73.53} \\ \hline
    EBM (800) & \textbf{83.10} & {51.37} & {64.20}  & {73.26} \\ \hline
   LBM & {82.69} & {48.38} & 60.11  & 70.88 \\ \hline
   CPM & {76.35} & {49.39} & 61.90  & 71.80 \\ \hline
   PM  & \textbf{84.0} & {47.34} & 61.30  & 71.60 \\ \hline   
    \multicolumn{5}{|c|}{Train: Pascal12 + ImageNet, Test: Pascal12} \\ \hline
      EBM (800)  & { 83.79} & {51.89} & {60.74} &  {75.39} \\ \hline
        \multicolumn{5}{|c|}{Train: Pascal12 + ImageNet, Test: Pascal12 + ImageNet} \\ 
        \hline
      EBM (800) & {92.83} & {67.26} &  {75.11} & 83.27 \\ 
      \hline
    \end{tabular}
          \label{T:CompSOAPascal}
\vspace{-4mm}
\end{table}

\subsection{Computational Analysis and Convergence}
\label{sec_comp_analysis_exp}

\begin{figure}[ht]
\centering
  \includegraphics[width=5.2in]{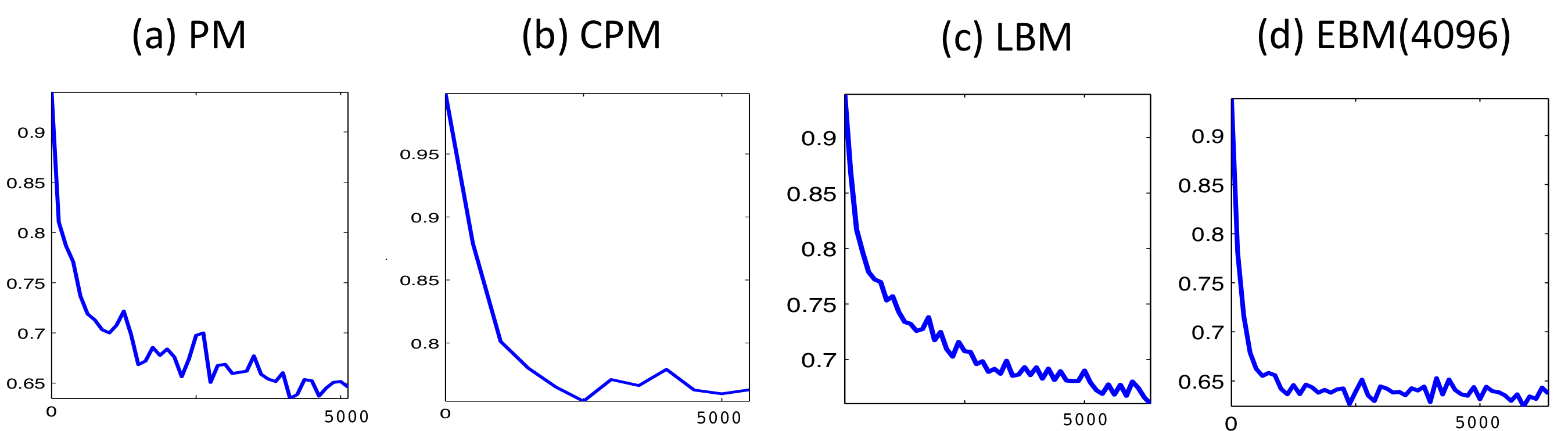}
  \caption{ Comparison of convergence between the models. On the left is the category error and on the right is the pose error, on the validation set, respectively for each model (a) to (d). The error is computed per batch during each iteration. CPM shows the error for the joint category and pose. It can be seen that EBM (4096) converges much faster than the others which is another benefit of early branching. This is despite have a lot more parameters than the other models. This indicates that each of the subnetworks of EBM are able to specialize in both categorization and pose estimation faster. Each iteration is computed on one batch of 100 training samples.} 
  \label{F:Convergence}
\end{figure}

We performed computational analysis on the convergence of the models and show that EBM converges substantially faster than all the other models. In Fig.~\ref{F:Convergence} we show the convergence rates of the proposed models. EBM here is the larger EBM (4096) network. Despite having many more parameters than most of the other models (about $\sim$112 million parameters compared to 60 million in the base network), EBM converges substantially faster than all the other models. This shows the ability of this particular network to specialize faster in the two tasks. The shared first five layers are able to build up the object-view manifolds, preserve them and enhance them in the pose subnetwork of the model, while the other subnetwork specializes in pose-invariant category recognition. 

\vspace{-2mm}
\section{Discussion}
\vspace{-2mm}
\label{S:Discussion}
Analysis of the layers of all the CNN models is shown in Fig. \ref{fig:SVMRidge_rgbd} to \ref{F:kNN2}. We provide quantitative results over two challenging datasets and summarize them in Tables \ref{T:CompCNN}, \ref{T:CompSOA} and \ref{T:CompSOAPascal}.


We compare our models with multiple baselines in Table \ref{T:CompCNN}: Linear SVM and Kernel Ridge-Regression on HOG descriptors \cite{Dalalcvpr05:HoG} as well as on features extracted from the best performing layers of the base network on each task. These baseline results were expected to be quite lower than our models' performance due to the lack of fine-tuning in the base model and due to the sharing of the network layers between the two tasks of category recognition and pose estimation.

Without fine-tuning the base network does not represent the object-view manifold well enough to estimate the pose efficiently. After fine-tuning on each of the respective datasets, we were able to achieve good category performance using the PM model. The downside of this model is its inability to perform robust pose estimation on the more challenging natural sparser-views of Pascal3D+. This is evident in the results shown in Table \ref{T:CompSOAPascal}, where PM achieves less pose accuracy. 
 
In Tables \ref{T:CompCNN} and \ref{T:CompSOAPascal}, we see again that CPM does worse than the other models in both datasets. This is more evident in the task of categorization, \eg, a drop of $\sim$7\% and 2\%-3\% in category and pose accuracy on Pascal3D+, respectively, and similarly $\sim$10\% and $\sim$6\% on the RGBD dataset. This motivates the need for branching in the networks and branching at the particular layer that better represents both category and pose. Interestingly, we found that CPM performs relatively better on Pascal3D+. We argue that the reason is that object poses in natural images are dominated by a smaller range of viewpoints and hence most of the pose-bins have  vanishing probability (easier to learn). In addition, Pascal3D+ has a smaller number of categories.



LBM performs relatively well on RGBD, but not on Pascal3D+. This can be attributed to the fact that RGBD has many more categories and is composed of images of objects under controlled settings and not \textit{in-the-wild} like in Pascal3D+. The images in the RGBD dataset are captured at dense views as the object rotates on a turn-table. This is why the pose information is more prevalent in the last layers. This is evident from the steep monotonically increasing curve of LBM in Fig. \ref{fig:SVMRidge_rgbd}-Left. This is not the case in Pascal3D+ where the increase is more steady and in fact there is a decrease after layer FC6 (see Fig. \ref{fig:SVMRidge_pascal}-Right).  




The reason why EBM performs better than PM even though its weights are randomly initialized is that PM's FC6 and FC7 layers in the pose-specific branch are initialized with category-specific weights from pre-training. This adversely affects pose estimation since it is a contradictory task that requires \textit{view-variant} representations and not view-invariant representations like that required in categorization. Therefore initializing FC6/FC7 by another network trained for a different task is not likely to help. We show that learning the convolutional filters jointly with categories help make them discriminative for both tasks and thus achieves a good accuracy on both tasks (see  Fig. \ref{fig:SVMRidge_rgbd} to \ref{F:kNN2} and Tables \ref{T:CompCNN}, \ref{T:CompSOA} and \ref{T:CompSOAPascal}). 

Comparing between EBM and LBM, we see that early branching is able to achieve a good balance between categorization and pose estimation by sharing the representations up to where we found the layer representation still capture pose information. We see this in Tables \ref{T:CompCNN}, \ref{T:CompSOAPascal} and Figures \ref{F:CatPoseModel0x} to \ref{F:kNN}, where better pose accuracy and slightly better categorization accuracy is achieved by EBM. We also see that in Figure \ref{F:kNN} that the object view-manifold collapses in the last two layers (one layer before LBM) and thus achieves better pose discrimination than LBM. 

The slight effect of decreasing the size of the layers in the pose subnetwork of EBM can be observed from the results in Table \ref{T:CompCNN} and \ref{T:CompSOAPascal}. 






\vspace{-2mm}
\section{Conclusion}
\vspace{-2mm}
\label{S:Conclusion}
This paper is the first exploration of using CNNs for object pose estimation. We present our analysis and comparison of CNN models with the goal of performing both efficient object categorization and pose estimation. Despite the dichotomy in categorization and pose estimation, we show how CNNs can be adapted, by novel means introduced in this paper, to simultaneously solve both tasks. We make key observations about the intrinsics of CNNs in their ability to represent pose. We quantitatively analyze the models on two large challenging datasets with extensive experiments and achieve better than state-of-the-art accuracy on both datasets.

%
%
 


\bibliography{ALL}
\bibliographystyle{iclr2016_workshop}

\begin{appendices}

\section{$k$-NN Results}
\label{app_KNN1}

\subsection{Pascal3D}
KNN figures for categorization on Pascal3D dataset.
\begin{figure}[ht]
\centering
  \includegraphics[width=5.2in,height=3.4in]{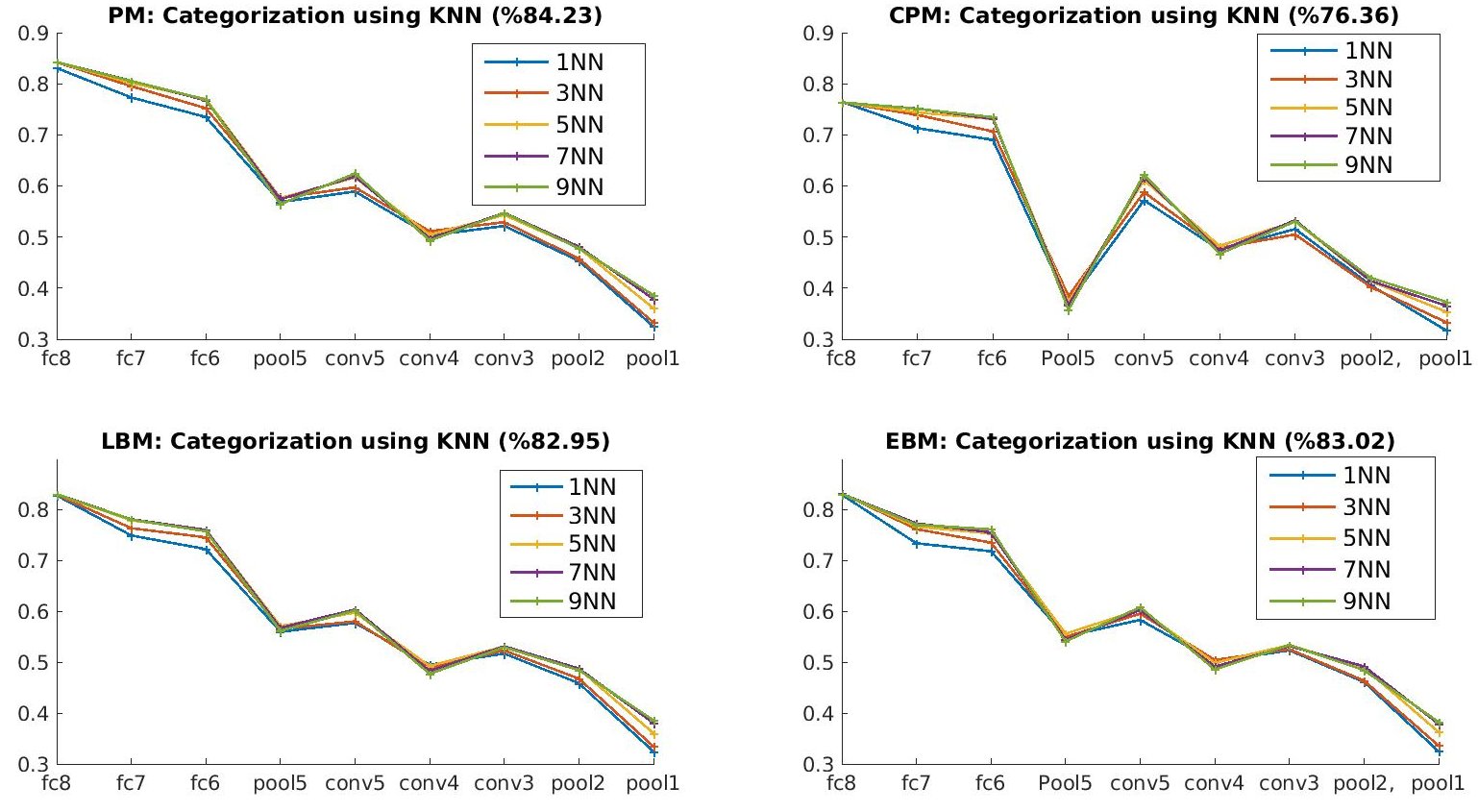}
   \caption{Comparison of the categorization at each layer of the CNN using $k$-NN with varying $k=\{1,3,5,7,9\}$ from top to bottom. This experiment was conducted on the PASCAL3D dataset categorization} 
  \label{F:kNN2}
\end{figure}
\clearpage
\subsection{RGBD Dataset}

It is clear how CPM is very unstable for dense poses that exist in RGBD dataset.

\begin{figure}[ht]
  \vspace{-5mm}
\centering
 \hspace{-5mm}  \includegraphics[width=6.2in,height=4.0in]{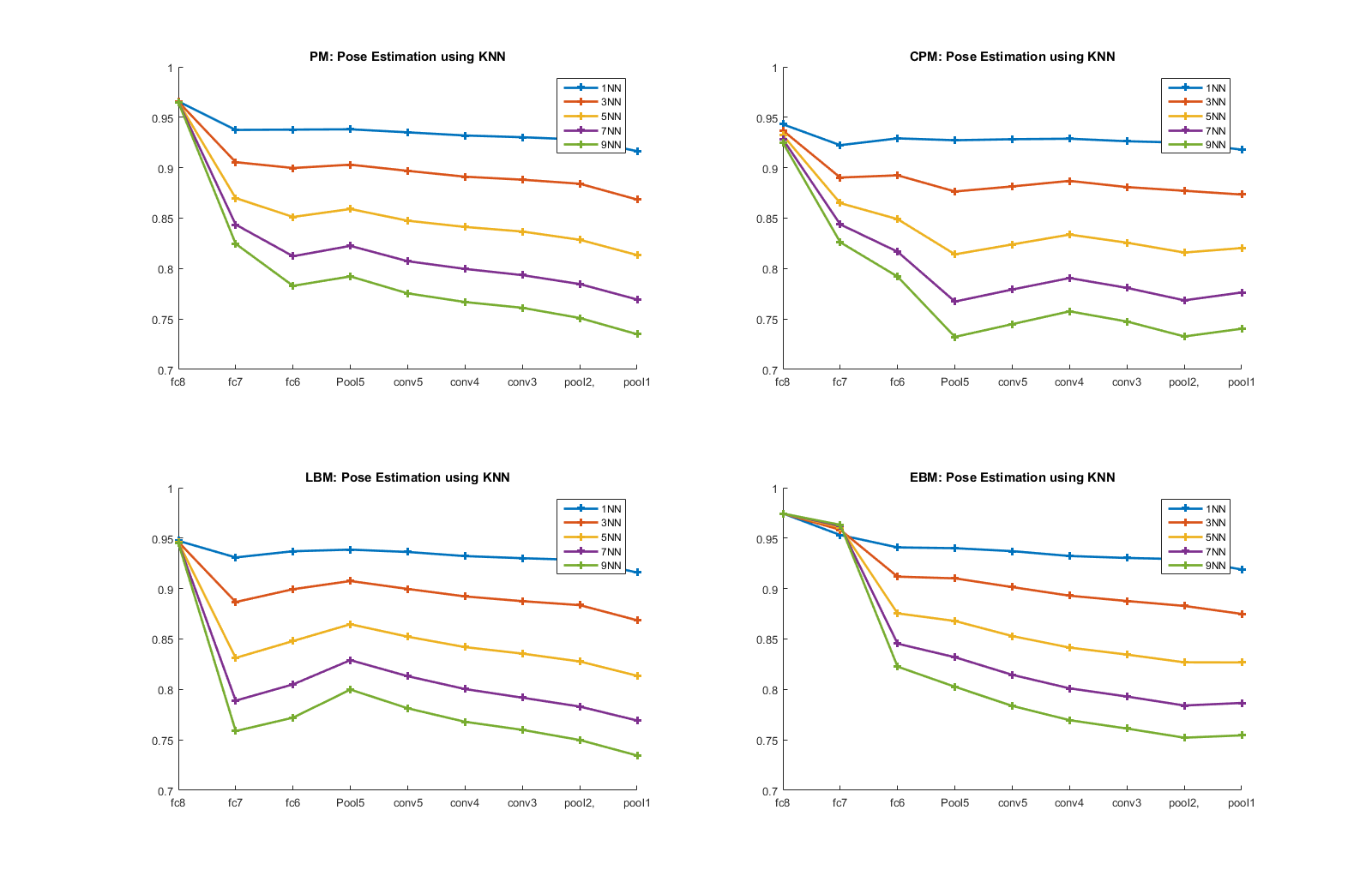}
   \vspace{-10mm}
   \caption{Comparison of the pose estimation at each layer of the CNN using $k$-NN with varying $k=\{1,3,5,7,9\}$ from top to bottom. This experiment was conducted on the RGBD dataset categorization (training points)} 
  \label{F:kNN2}
  \vspace{-5mm}
\end{figure}

\begin{figure}[ht]
  \vspace{-5mm}
\centering
\hspace{-5mm}  \includegraphics[width=6.2in,height=4.0in]{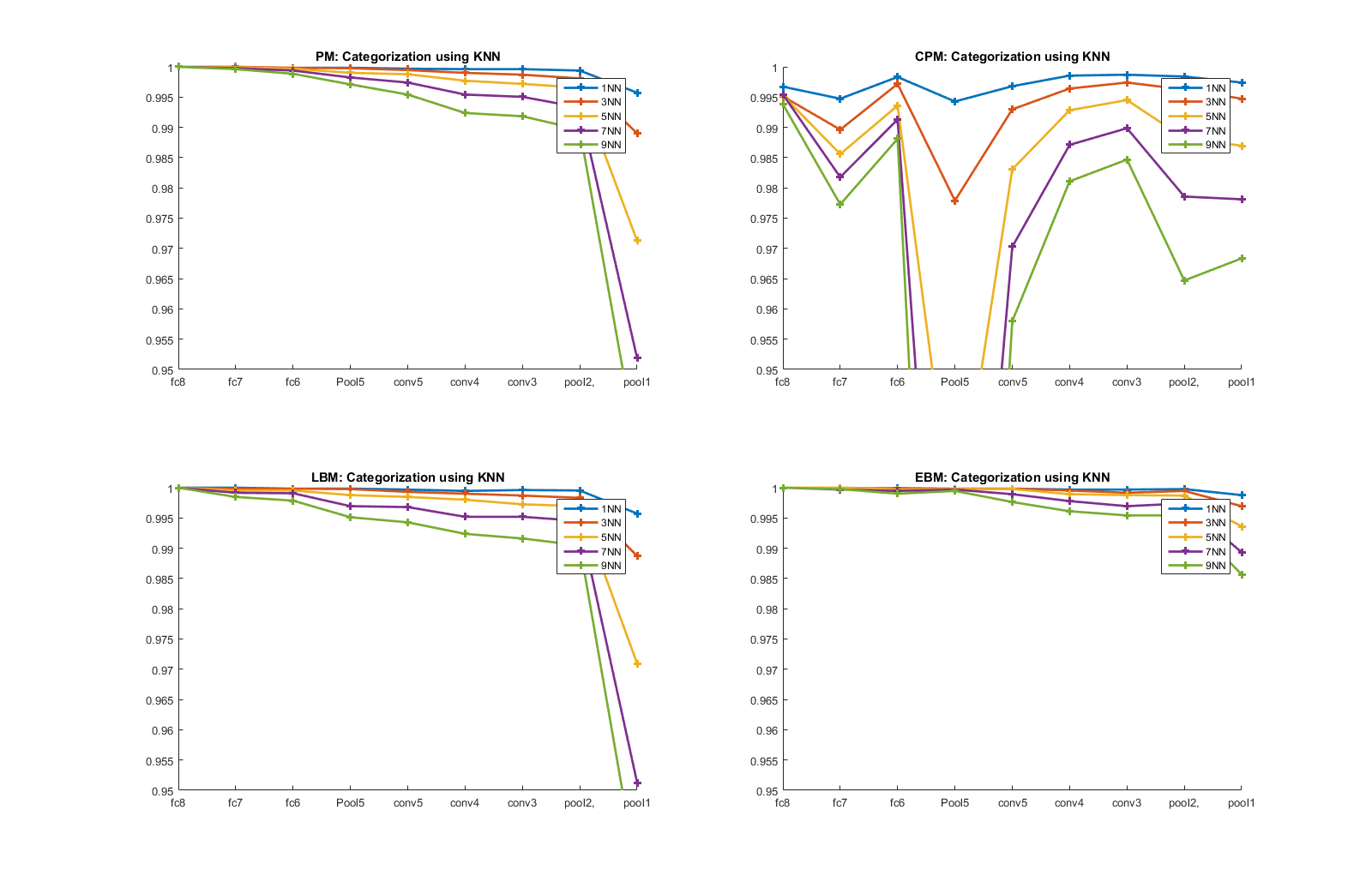}
  \vspace{-10mm}
   \caption{Comparison of the categorization  at each layer of the CNN using $k$-NN with varying $k=\{1,3,5,7,9\}$ from top to bottom. This experiment was conducted on the RGBD dataset categorization (training points)} 
  \label{F:kNN2}
    \vspace{-10mm}
\end{figure}

\clearpage

\section{Local Pose Measurement Analysis on RGBD Dataset}
\label{sec:localmeas}
We applied three  local measurement analysis proposed in~\citep{bakry2015digging} to analyze features against dense poses. For more details about the description of these measurements, please refer to~\citep{bakry2015digging}. The main property that these measurements quantified is how these representations align with the the circle manifold that represent the pose of the categories. 

 All the figures shows that EBM achieves the best behavior in untangling both the categorization and the pose branches. It is clear that CPM behaves the worst for pose estimation as we argued in the paper for several reasons. 

\begin{enumerate}
\item \textbf{Z-EffectiveSV 90 (The lower the better)}
\item \textbf{ TPS-RCond-CF-poly (The higher the better)}"
\item \textbf{Nuclear Norm (The higher the better)}
\item \textbf{KPLS-Kernel Regression Error (The lower the better)}
\end{enumerate}

\begin{figure}[ht]
\centering
 \includegraphics[width=6.2in,height=3.0in]{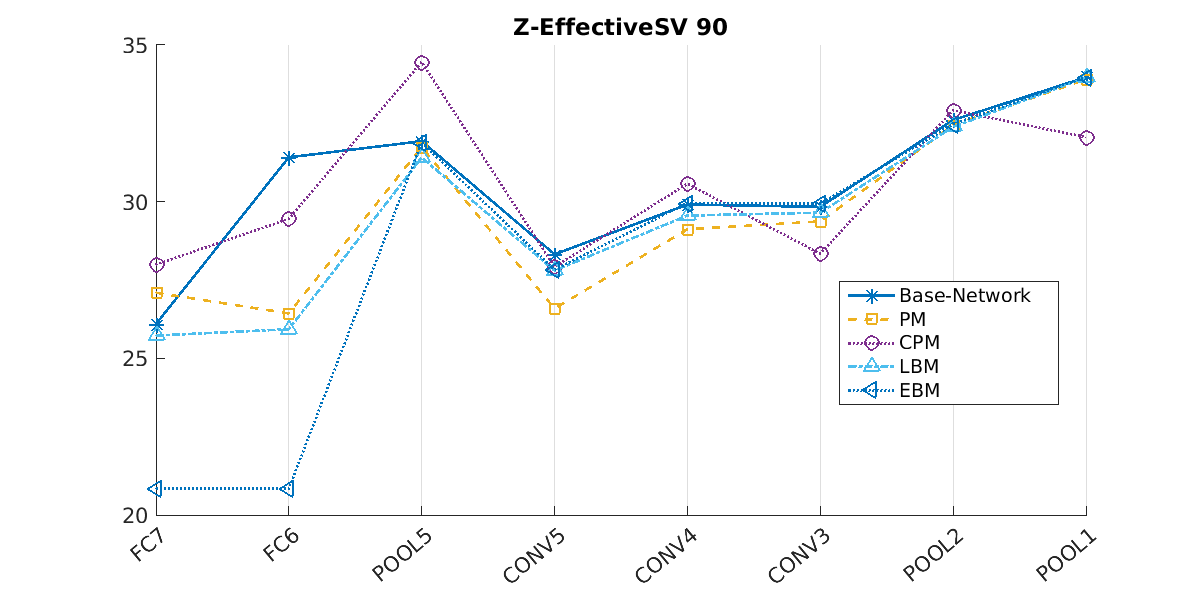}
   \caption{Comparison of the pose estimation  at each layer of the CNN using ``Effectibe SV 90\%''}
  \label{F:local_SV90}
\end{figure}

\begin{figure}[ht]
\centering \includegraphics[width=6.2in,height=3.0in]{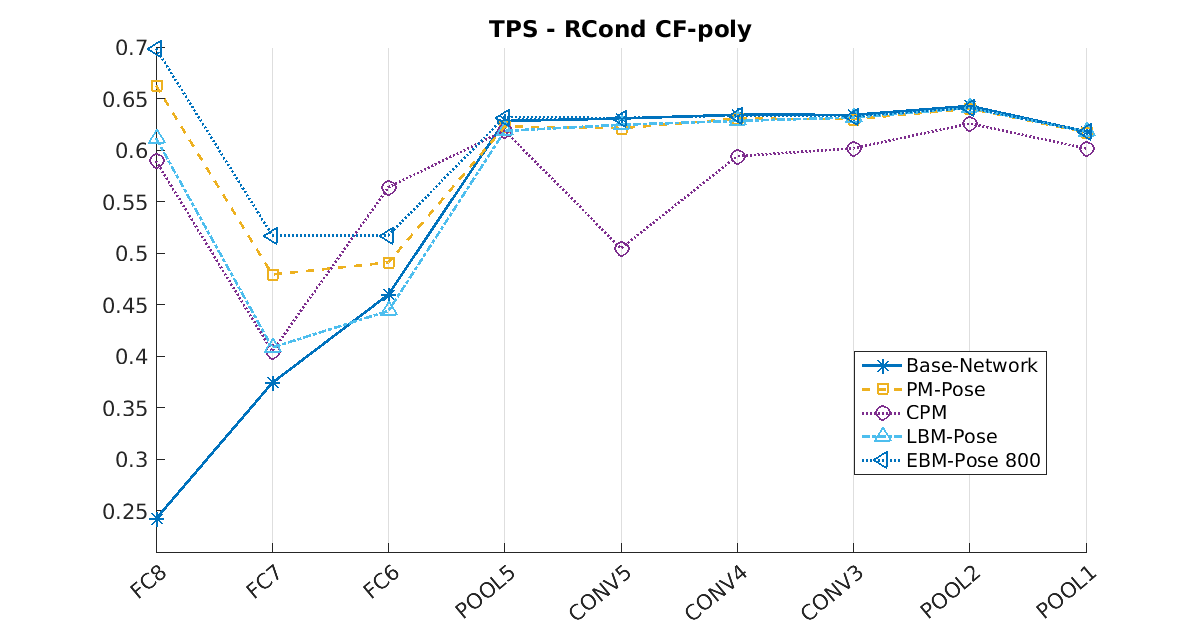}
   \caption{Comparison of the pose estimation  at each layer of the CNN using ``TPS-RCond (polynomal)['' measurement}
  \label{F:local_TPS}
\end{figure}

\begin{figure}[ht]
\centering
\hspace{-5mm}  \includegraphics[width=6.2in,height=3.0in]{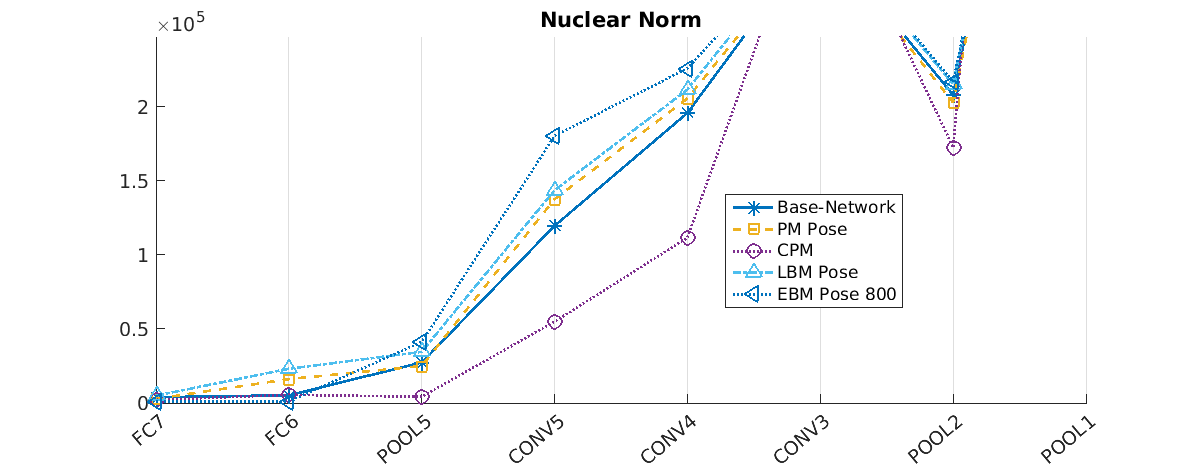}
   \caption{Comparison of the pose estimation  at each layer of the CNN using ``Nuclear Norm'' measurement (FC8 to Pool5)}
  \label{F:local_NNorm1}
\end{figure}

\begin{figure}[ht]
\centering
 \includegraphics[width=6.2in,height=3.0in]{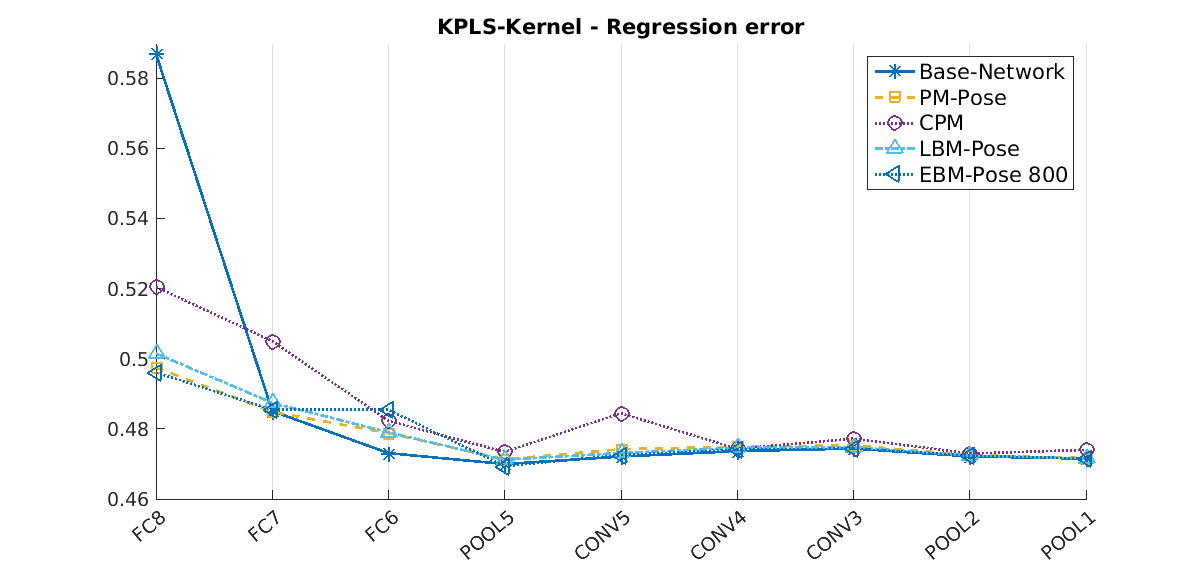}
   \caption{Comparison of the pose estimation  at each layer of the CNN using KPLS Kernel Regression Error measurement}
  \label{F:local_KPLS}
\end{figure}

\clearpage

\section{Computational Analysis and Convergence(More Details)}



\begin{figure*}[ht!]
  \centering
  \includegraphics[width=0.9\textwidth, height=0.3\textwidth]{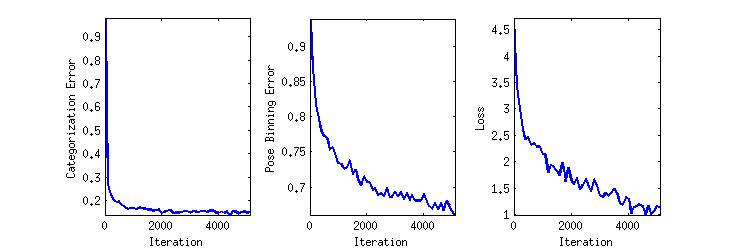}
  \caption{LBM Training (Categorization Error on the left, Pose Binning Error in the middle, Loss on  the right)}
  \label{figmodel4train}
\end{figure*}

\begin{figure*}[ht!]
  \centering
  \includegraphics[width=0.9\textwidth, height=0.3\textwidth]{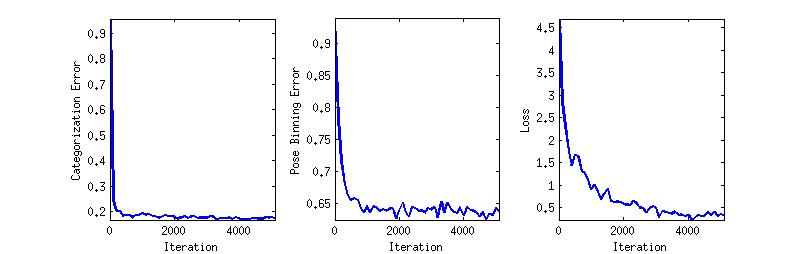}
  \caption{EBM Training (Categorization Error on the left, Pose Binning Error in the middle, Loss on the right)}
  \label{figmodel5train}
\end{figure*}

The following figures show the loss/validation curves for the trained CNNs. The loss is shown per the batch being processed at each iteration(one training batch/iteration). The interesting behavior we notice in all the networks is that the categorization part converges very quickly, while the pose part takes sometime to converge. This is since, in all networks, the layers were initialized with the ImageNet categorization CNN. For the pose part, initialization for a categorization network might not be helpful especially for the top layers (e.g. FC6,FC7, and FC8), since they were trained for a different purpose that might be conflicting.   Furthermore, training on a joint loss as in EBM and LBM positively affects the convergence  as can be seen in figures~\ref{figmodel4train} and  ~\ref{figmodel5train}. It is not hard to see that the Early Branching reduced the validation pose error significantly faster compared to the remaining models despite having much more parameters than many of the other 
models. Refer to section 4 for the number of parameters.

\begin{figure*}[ht!]
  \centering
  \includegraphics[width=0.6\textwidth, height=0.3\textwidth]{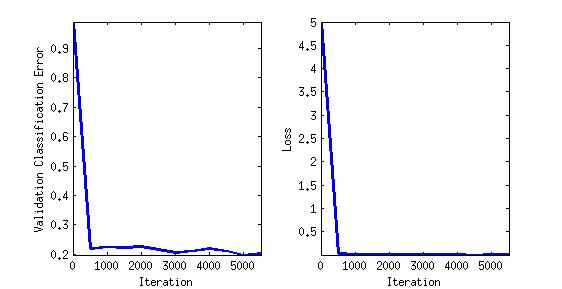}
  \caption{PM Category CNN Training}
  \label{figmodel1cattrain}
\end{figure*}

\begin{figure*}[ht!]
  \centering
  \includegraphics[width=0.6\textwidth, height=0.3\textwidth]{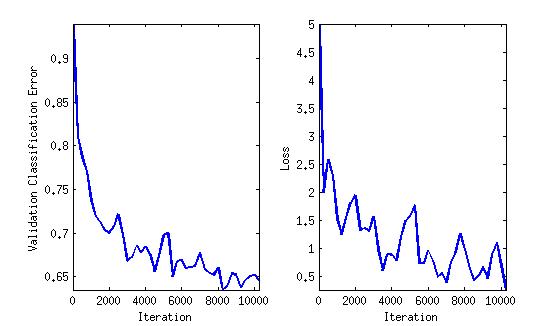}
  \caption{PM Pose CNN Training}
  \label{figmodel1posetrain}
\end{figure*}

%
%
%
%

\begin{figure*}[ht!]
  \centering
  \includegraphics[width=0.6\textwidth, height=0.3\textwidth]{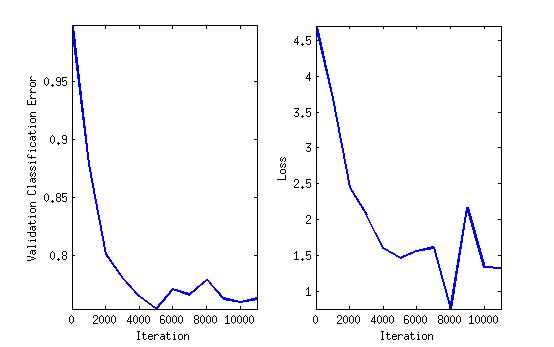}
  \caption{CPM Training Error (this is the error of both classifying the correct category in the correct pose bin)}
  \label{figmodel3train}
\end{figure*}

\clearpage

\section{Error Metrics}
\label{app_metrics}

The two metrics $<22.5$ and $<45$ used to evaluate the performance of pose estimation are the percentages of test samples that satisfy $AE<22.5^{\circ}$ and $AE<45^{\circ}$, respectively, where the Absolute Error (AE) is $AE= | Estimated Angle - Ground Truth|$). The AAAI pose accuracy (used extensively in the previous work we compare with) is equal to $1 - [min{(|\theta_i - \theta_j|,2\pi - |\theta_i - \theta_j|)} / \pi]$.

\section{Training Parameters}
\label{app_training}
The base learning rate is assigned $0.5 \times 10^{-3}$. For fine-tuning, the learning rate of the randomly initialized parameters (\emph{e.g.} FC8 parameters in PM) are assigned to be ten times higher than the learning rate of the parameters initialized from the pretrained CNN (\emph{e.g.} Conv1 to Pool5 in all the models). The decay of the learning rate $\gamma$ is 0.1. While training our CNNs, we drop the learning rate by a factor of $\gamma$ every 5000 iterations. The momentum and the weight decay were assigned to 0.9 and 0.0001 respectively. Training images are randomly shuffled before feeding the CNN for training. The training batch size was 100 images. 

\section{Initialization and Models' Parameters}
The ImageNet CNN used in our paper (AlexNet) [16] has $\sim$60 million parameters. In this section, we present how all the models were initialized in our experiments. Then, we analyze the number of parameters in the model for each of RGBD and Pascal3D datasets. 

\subsection{Initialization}
\subsubsection{EBM Model}

In EBM, we initialize all the convolutional layers by the convolution layer parameters of AlexNet. We initialize FC6 and FC7 of the category branch by the parameters of AlexNet model. The remaining layers were initialized randomly (\ie FC6, FC7, and  FC8 of the pose branch subnetwork, and FC8 of the category Branch subnetwork).

\subsubsection{LBM Model}

For LBM, we initialize all the layers by the pretrained AlexNet model for the convolution layers, FC6, and FC7. FC8 weights are initialized randomly.

\subsubsection{CPM Model}
We initialize all the layers by the pretrained AlexNet model for the convolution layers, FC6, and FC7. We initialized FC8 parameters randomly. 

\subsubsection{PM Model}
Since there are two separate models, one for category and one for Pose. We initialize all the layers by the pretrained AlexNet model for the convolution layers, FC6, and FC7. We initialized FC8 parameters randomly. 

\subsection{Number of Model Parameters for RGBD Dataset}

\subsubsection{EBM Model}
 EBM Model has 111,654,944 parameters. These are as follows starting from the input layer (number of filters x filter width x filter height x number of channels): 96x11x11x3 + 256x5x5x48 + 384x3x3x256 + 384x3x3x192 + 256x3x3x192) +  Pool5-FC6 (pose) 9216x4096,  FC6(pose)-FC7(pose) 4096x4096,  FC7(pose)-FC8(pose) 4096x16
 Pool5-FC6(category) 9216x4096,  FC6(category)-FC7(category) 4096x4096,  FC7(category)-FC8(category) 4096x51.
 
 \subsubsection{LBM Model}
LBM  has  57,133,088 params =
 convolution-layers' parameters (96x11x11x3 + 256x5x5x48 + 384x3x3x256 + 384x3x3x192 + 256x3x3x192) + 9216x4096 + 4096x4096 + (fc8-cat) 4096x51 + (fc8-pose) 4096x16. These are organized as (number of filters x filter width x filter height x number of channels) and starting from the input layer.

\subsubsection{CPM Model}
CPM  has  60,200,992 params =
 convolution-layers' parameters (96x11x11x3 + 256x5x5x48 + 384x3x3x256 + 384x3x3x192 + 256x3x3x192) + 9216x4096 + 4096x4096 + (fc8-cat and pose) 4096x51x16. These are organized as (number of filters x filter width x filter height x number of channels) and starting from the input layer.
 
 \subsubsection{PM Model}
PM   has  113,991,744 parameters  (56,924,192 for pose and 57,067,552 for category). These are shown below as (number of filters x filter width x filter height x number of channels) and starting from the input layer.

\textbf{PM Model Pose Parameters: }
The 56,924,192  pose parameters comes from  convolution-layers' parameters (96x11x11x3+ 256x5x5x48 + 384x3x3x256 + 384x3x3x192 + 256x3x3x192) +  fully connected layers' parameters (9216x4096 + 4096x4096 +  4096x16).

\textbf{PM Model Category Parameters: } The 57,067,552  category parameters comes from   convolution-layers' parameters  (96x11x11x3+ 256x5x548 + 384x3x3x256 + 384x3x3x192 + 256x3x3x192) + fully connected layers' parameters (9216x4096, + 4096x4096 + 4096x51).

\subsection{Number of Model Parameters for Pascal3D Dataset}

\subsubsection{EBM Model}
EBM Model 111,495,200 params =
convolution layers' parameters (96x11x11x3 + 256x5x5x48 + 384x3x3x256 + 384x3x3x192 + 256x3x3x192) +  Pool5-FC6 (pose) 9216x4096,  FC6(pose)-FC7(pose) 4096x4096,  FC7(pose)-FC8(pose) 4096x16
 Pool5-FC6(category) 9216x4096,  FC6(category)-FC7(category) 4096x4096,  FC7(pose)-FC8(pos) 4096x16. These are organized as (number of filters x filter width x filter height x number of channels) and starting from the input layer.

\subsubsection{LBM Model}
LBM  has 56,969,248 params =
convolution layers' parameters (96x11x11x3 + 256x5x5x48 + 384x3x3x256 + 384x3x3x192+ 256x3x3x192)+ 9216x4096 + 4096x4096 + (fc8-cat) 4096x11 + (fc8-pose) 4096x16. These are organized as (number of filters x filter width x filter height x number of channels) and starting from the input layer.

\subsubsection{CPM Model}
CPM  has  57,579,552 params =
convolution layers' parameters (96x11x11x3 + 256x5x5x48 + 384x3x3x256 + 384x3x3x192 + 256x3x3x192) + 9216x4096 + 4096x4096 + (fc8-cat and pose) 4096x11x16. These are organized as (number of filters x filter width x filter height x number of channels) and starting from the input layer.

\subsubsection{PM Model}
PM   has  113,827,904 parameters  (56,924,192 for pose and 56,903,712 for category. These are shown below as (number of filters x filter width x filter height x number of channels) and starting from the input layer.

\textbf{PM Model Pose Parameters: }
The 56,924,192  pose parameters comes from convolution layers' parameters (96x11x11x3+ 256x5x5x48 + 384x3x3x256 + 384x3x3x192 + 256x3x3x192) + fully connected layers' parameters  (9216x4096, + 4096x4096 + 4096x16).

\textbf{PM Model Category Parameters: }
 The 56,903,712  category parameters comes from convolution layers' parameters (96x11x11x3 + 256x5x5x48 + 384x3x3x256 + 384x3x3x192 + 256x3x3x192) + fully connected layers' parameters (9216x4096, + 4096x4096 + 4096x11).

\section{Effect of Loss Function Weights on EBM model}
\label{app_weights}

We found that changing the weights for EBM slightly affected the performance; see table~\ref{tbl_bh}. Our intuition behind this behavior is that EBM splits into separate parameters starting from Pool5, which makes each of the pose and the category have some independent parameters (FC6,FC7,FC8) in addition to the shared parameters (Conv1 to Pool5).  

\begin{table}[htb!]
\caption{Effect of $\lambda_1$ and $\lambda_2$  for EBM}
\centering \footnotesize
\begin{tabular}{|p{2.5cm}|p{2.6cm}|p{1.5cm}|}
\hline
Parameters & Categorization \% & Pose \% \\
\hline 
$\lambda_1=1$  $\lambda_2=1$ & \textbf{89.94 }& \textbf{ 82.00}\\ 
\hline 
$\lambda_1=1$  $\lambda_2=2$  &    89.39  & 81.80\\ 
\hline 
$\lambda_1=2$  $\lambda_2=1$ &  89.25 & 81.89\\ 
\hline 
\end{tabular} 
\label{tbl_bh}
\end{table}

Since $\lambda_1=1$ , $\lambda_2=1$ is slightly better than others, we performed all of our Model 5 experiments in the paper with this setting.

\end{appendices}


\end{document}